\documentclass{article}

\usepackage{arxiv}

\usepackage{researchpack}
\usepackage{subfigure}
\usepackage{natbib}
\usepackage{hyperref}
\usepackage[capitalise,nameinlink]{cleveref}
\usepackage{xspace}
\usepackage{enumitem}
\usepackage{annotate-equations}
\usepackage{rotating}
\usepackage{threeparttablex}
\usepackage{tcolorbox}
\usepackage{pifont}
\usepackage{wrapfig}
\usepackage{tikz}
\usepackage{enumitem}
\usepackage{array}
\usepackage{centernot}
\usepackage{adjustbox}

\usetikzlibrary{positioning,shapes,arrows,calc}

\hypersetup{
    colorlinks=true,
    citecolor=teal,
    linkcolor=violet,
    urlcolor=teal,
    pdftitle={DeepDFA: Injecting Temporal Logic in Deep Learning for Sequential Subsymbolic Applications},
    pdfauthor={Elena Umili, Francesco Argenziano, Roberto Capobianco},
    pdfkeywords={Neuro-Symbolic AI, Symbol Grounding},
}

\newcommand*{\colorboxed}{}
\def\colorboxed#1#{%
  \colorboxedAux{#1}%
}
\newcommand*{\colorboxedAux}[3]{%
  \begingroup
    \colorlet{cb@saved}{.}%
    \color#1{#2}%
    \boxed{%
      \color{cb@saved}%
      #3%
    }%
  \endgroup
}

\begin{document}

\title{DeepDFA:\\ Injecting Temporal Logic in Deep Learning for Sequential Subsymbolic Applications}

\author{
    Elena Umili\thanks{
    $^*$ Corresponding author
    }\\
    Sapienza University of Rome\\Italy\\
    \texttt{surname@diag.uniroma1.it}\\
    \And
    Francesco Argenziano\\
    Sapienza University of Rome\\Italy\\
    \texttt{surname@diag.uniroma1.it}\\
    \And
    Roberto Capobianco\\
    Sony AI\\Switzerland\\
    \texttt{name.surname@sony.com}\\
}

\setshorttitle{DeepDFA}
\setshortauthors{Umili et al.}

\maketitle

\begin{abstract}
    Integrating logical knowledge into deep neural network training is still a hard challenge, especially for sequential or temporally extended domains involving subsymbolic observations. To address this problem, we propose DeepDFA, a neurosymbolic framework that integrates high-level temporal logic—expressed as Deterministic Finite Automata (DFA) or Moore Machines—into neural architectures. DeepDFA models temporal rules as continuous, differentiable layers, enabling symbolic knowledge injection into subsymbolic domains. We demonstrate how DeepDFA can be used in two key settings: (i) static image sequence classification, and (ii) policy learning in interactive non-Markovian environments. Across extensive experiments, DeepDFA outperforms traditional deep learning models (e.g., LSTMs, GRUs, Transformers) and novel neuro-symbolic systems, achieving state-of-the-art results in temporal knowledge integration. These results highlight the potential of DeepDFA to bridge subsymbolic learning and symbolic reasoning in sequential tasks.
\end{abstract}

\maketitle

\section{Introduction}
Artificial Intelligence (AI) systems are increasingly deployed in complex scenarios that demand both perception and reasoning capabilities.
While deep learning \cite{lecun2015deep} and deep reinforcement learning \cite{mnih2015human, silver2016mastering} have achieved remarkable success on subsymbolic tasks involving raw sensory data, these models often fail to satisfy even basic logical constraints \cite{roadR_giunchiglia, STLnet}, derived from commonsense or domain-specific knowledge.
This shortcoming arises because such models are typically trained solely on data, leaving the integration of logical knowledge into learning pipelines an open and challenging problem.
As a result, in many real-world settings where both data and symbolic knowledge are available, the latter is often underutilized.

To address this issue, Neuro-Symbolic (NeSy) AI has emerged as a promising paradigm for integrating raw data with structured symbolic knowledge during the training of neural networks, helping to bridge the well-known Symbol Grounding Problem \cite{symbol_grounding_problem}.
Much of the existing NeSy literature focuses on injecting logical constraints into complex subsymbolic tasks \cite{darwiche2011, DILIGENTI2017, xu2018, LTN, DeepProbLog, neurasp, deepstochlog}.
However, these methods are rarely designed for sequential tasks where logical rules unfold over time, and are best expressed using formalisms such as Regular Expressions, Deterministic Finite Automata (DFAs), or Linear Temporal Logic over finite traces (LTLf).
Yet, temporal rules are widely used across many domains—ranging from robotics \cite{suriani_ltl_soccer} to healthcare \cite{patrizi_health} to business process management \cite{LTLfBPM}—for tasks involving sequential decision-making and reasoning, including planning \cite{planningltl}, reinforcement learning \cite{hidden_triggers}, and reactive synthesis \cite{reactive_synthesis_ltl}.
Moreover, the few approaches capable of incorporating temporal knowledge \cite{Umili2023KR, NeSyA} are generally applied only in static sequence classification, with little exploration of their applicability in interactive, decision-making environments.

In this paper, we propose DeepDFA, a general framework for integrating temporal logical knowledge into neural systems. DeepDFA is a continuous and differentiable logic layer that can represent temporal rules in the form of Deterministic Finite Automata (DFA) or Moore Machines. Conceptually, it operates as a hybrid between a Recurrent Neural Network and a symbolic automaton. Built on the theoretical foundations of Probabilistic Finite Automata (PFA), DeepDFA enables symbolic knowledge to be encoded as neural components that are fully compatible with gradient-based optimization. This design allows temporal knowledge to be injected by: (i) first encoding temporal knowledge as fixed network parameters, and (ii) defining loss functions that encourage or discourage the satisfaction of such knowledge, depending on the context. 

While this provides a principled approach to integrating temporal knowledge across a wide range of applications, in this work we focus on subsymbolic domains that require a perception—or grounding—phase for logical symbols. We assume that the grounding function is not available as prior knowledge and must instead be learned by the system exploiting both the data and the logical knowledge available. This assumption reflects a realistic setting: while symbolic temporal knowledge can often be shared across different domains, the perceptual grounding is inherently domain-specific and should not be hard-coded into the system.
To this end, we demonstrate how DeepDFA can be seamlessly integrated into subsymbolic sequential pipelines involving perception, enabling its application in two key settings:
(i) static sequence classification from raw sensory inputs, such as activity recognition in videos; and
(ii) non-Markovian reinforcement learning in non-symbolic environments, where decision-making depends on temporal context \cite{neural_reward_machine_ecai}.

Experimental results across multiple benchmarks show that our method outperforms both deep learning models for sequential data (e.g., RNNs, Transformers) and novel neurosymbolic systems (FuzzyDFA \cite{Umili2023KR}, NeSyA \cite{NeSyA}), achieving state-of-the-art results in temporal knowledge integration. These results highlight the potential of DeepDFA to bridge subsymbolic learning and symbolic reasoning in sequential tasks.

The remainder of this paper is organized as follows: Section 2 reviews related work. Section 3 provides background on temporal logics and automata. In Section 4, we introduce the DeepDFA architecture. Section 5 illustrates how DeepDFA is used for knowledge integration in classification and RL. Section 6 presents our experimental evaluation, and Section 7 concludes with final remarks and directions for future work.

\section{Related Works}
\paragraph{Knowledge Integration in Subsymbolic Applications}
Much prior work approaches the problem of injecting prior logical knowledge into deep learning methods for subsymbolic applications \cite{ xu2018, LTN, DeepProbLog, Umili2023KR, abduction2019, huang_abduction_2021, Tsamoura_abduction_2021}. An open challenge in these kinds of domains is to address the so-called symbol grounding (SG) problem, i.e. the problem of associating symbols with abstract concepts without explicit supervision \cite{symbol_grounding_problem}. To this end, external knowledge can be used to weakly supervise the perception function of symbols, we refer to this practice as semi-supervised symbol-grounding (SSSG). SSSG is tackled mainly with two families of methods.
The first approach \cite{ xu2018, LTN, DeepProbLog, Umili2023KR} consists in embedding a continuous relaxation of the available knowledge in a larger neural network system, 
and training the system so to align the outputs of the relaxation with the high-level labels.
The second approach \cite{abduction2019, huang_abduction_2021, Tsamoura_abduction_2021} instead maintains a crisp boolean representation of the logical knowledge and uses a process of logic \textit{abduction} to correct the current SG function, that is periodically retrained with the corrections.
Much work in this area does not take into account temporal aspects, and only few works \cite{DonadelloBPM24, NeSyA, Umili2023KR} focus on integrating knowledge for \textit{sequential} and temporal problems. In particular,
Donadello et al. \cite{DonadelloBPM24} design a fuzzy counterpart of LTLf, to perform conformance check on fuzzy-grounded values. However their method is not differentiable and therefore not easily integrable with deep learning model training.
FuzzyDFA \cite{Umili2023KR} and NeSyA \cite{NeSyA} are novel continuous and differentiable relaxations of DFAs, using respectively Fuzzy and probabilistic semantics.
Our approach is similar to these two, but is based on probabilistic Moore Machine and is more versatile than FuzzyDFA and NeSyA, as it can be used for both \textit{injecting} prior temporal knowledge and training it from data. Furthermore in \cite{Umili2023KR} and \cite{NeSyA} the authors provide an integration of their framework only in static sequence classification problems, while we provide in this paper a principled method for injecting temporal knowledge not only in static classification, but also in interactive Reinforcement Learning domains.

\paragraph{Non-Markovian Reinforcement Learning with Temporal Specifications}
Temporal logic formalisms are widely used in Reinforcement Learning (RL) to specify non-Markovian tasks \cite{littman2017}. 
The large majority of works assumes the boolean symbols used in the formulation of the task are \textit{perfectly} observable in the environment \cite{reward-machine-sheila, restr_bolts, reward_machine_learning_1, reward_machine_learning_2, reward_machine_learning_3, subgoalAutomaton}. For this reason they are applicable only in symbolic-state environments or when a perfect mapping between the environment state and a symbolic interpretation is known, also called labeled MDP \cite{continuous_restr_bolt}.
Many works assume to know an \textit{imperfect} SG function for the task \cite{noisy_symbols_2020, noisy_symbols_2022, noisy_symbols_2024_sheila}. Namely, a function that sometimes makes mistakes in predicting symbols from states or predicts a set of probabilistic \textit{beliefs} over the symbol set instead of a boolean interpretation. These works represent a step towards integration with non-symbolic domains. However, they do not address the problem of learning the SG function, but only how to manage to use a pre-trained imperfect symbol grounder.
Only one work in the literature assumes the same setting of ours \cite{ltl_no_grounding}, namely, that the agent observes sequences of non-symbolic states and rewards, and is aware of the LTL formula describing the task but without knowing the meaning of the symbols in the formula. This work employs a neural network composed of different modules, each representing a propositional symbol or a logical or temporal operator, which are interconnected in a tree-like structure, respecting the syntactic tree of the formula. This network learns a representation of states, that can be easily transferred to different LTL tasks in the same environment.
However, the key distinctions between our and their work are: (i) Kuo et al. \cite{ltl_no_grounding} learn a subsymbolic uninterpretable representation, while we learn \textit{precisely} a mapping between states and symbols; (ii) their method provides benefits only in a multitask setting and is unable to expedite learning on a single task, while ours learns and leverages the symbol grounding in the individual task.

\section{Background: Specifying Temporal Patterns with Logic}
Various formalisms exist in the literature to formally and unambiguously specify constraints and patterns over temporal traces. Here, we introduce some notation and give background knowledge on formal languages for time sequences. Finally, we will describe their use in the context of non-Markovian Reinforcement Learning.

\subsection{Notation}
In this work, we consider \textit{sequential} data of various types, including both symbolic and subsymbolic representations. Symbolic sequences are also called \textit{traces}. Each element in a trace is a symbol $\sigma$ drawn from a finite alphabet $\Sigma$.

We denote sequences using bold notation. For example, $\boldsymbol{\sigma} = (\sigma^{(1)}, \sigma^{(2)}, \ldots, \sigma^{(T)})$ represents a trace of length $T$. Each symbolic variable in the sequence can be grounded either categorically or probabilistically.

In the case of categorical grounding, each element of the trace is assigned a symbol from $\Sigma$, denoted simply as $\sigma^{(i)}$. In the case of probabilistic grounding, each symbolic variable is associated with a probability distribution over $\Sigma$, represented as a vector $\tilde{\sigma}^{(i)} \in \Delta(\Sigma)$, where $\Delta(\Sigma)$ denotes the probability simplex defined as
\[
\Delta(\Sigma) = \left\{ \tilde{\sigma} \in \mathbb{R}^{|\Sigma|} \,\middle|\, \tilde{\sigma}_j \geq 0,\ \sum_{j=1}^{|\Sigma|} \tilde{\sigma}_j = 1 \right\}.
\]

We use superscripts to indicate time steps in the sequence and subscripts to denote vector components. For instance, $\tilde{\sigma}^{(i)}_j$ denotes the $j$-th component of the probabilistic grounding of $\sigma$ at time step $i$.

Accordingly, we distinguish between categorically grounded and probabilistically grounded sequences using the tilde notation: $\boldsymbol{\sigma}$ denotes a sequence of categorical assignments (i.e., discrete symbols), while $\boldsymbol{\tilde{\sigma}}$ denotes a sequence of soft assignments (i.e., probability vectors).

\begin{figure*}[t]
    \centering
    \subfigure[]{
    \includegraphics[width=0.3\textwidth]{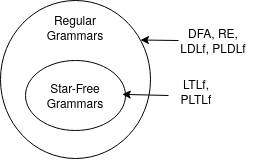}}\quad
    \subfigure[]{
    \includegraphics[width=0.22\textwidth]{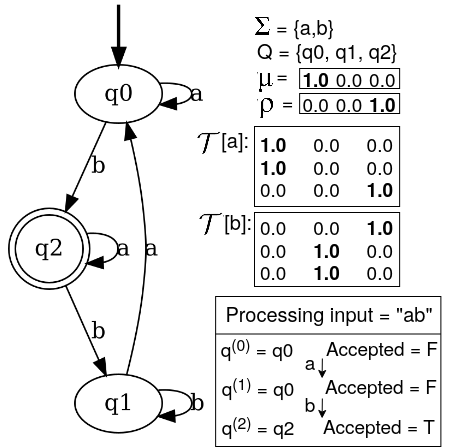}}\quad
    \subfigure[]{
    \includegraphics[width=0.37\textwidth]{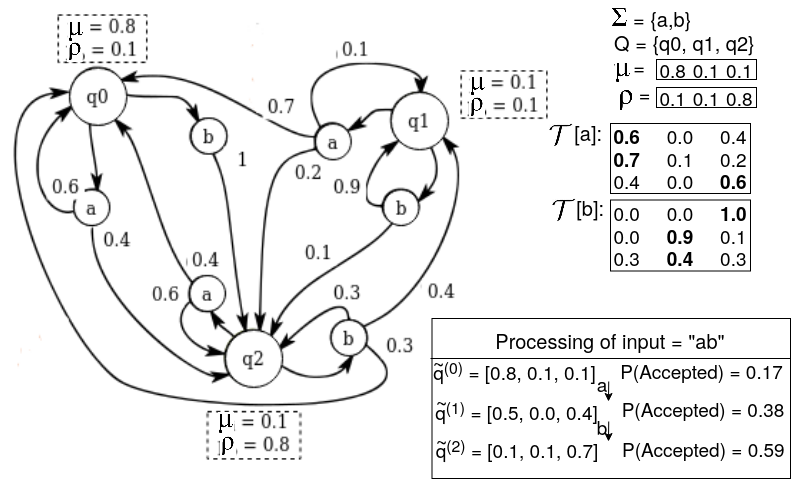}}

    \caption{a) Expressivity of different temporal formalisms. b-c) An example of a PFA (b) and a DFA (c) with three states and two symbols. For both we show: (i) the graph describing the automaton, (i) its equivalent representation in matrix form, and (ii) produced states and output probabilities while processing the string "ab".}
    \label{fig:PFAvsDFA}
\end{figure*}

\subsection{Deterministic Finite Automata and Moore Machines} \label{sec:DFA}
A Deterministic Finite Automaton (DFA) $A$ is a tuple $(\Sigma, Q, \delta, q_0,  F)$, where $\Sigma$ is the finite input alphabet, $Q$ is a finite set of states, $q_0 \in Q$ is the initial state, $\delta: Q \times \Sigma \to Q $ is the transition function, and $F \subseteq Q$ is the set of final states.
Let $\Sigma^*$ be the set of all finite strings over $\Sigma$, and $\epsilon$ the empty string.
The transition functions over strings $\delta^*:Q \times \Sigma^* \rightarrow Q$ are defined recursively as
\begin{equation}
\begin{array}{l}
    \delta^*(q,\boldsymbol{\epsilon}) = q \\
    \delta^*(q, a\boldsymbol{\sigma}) = \delta^*(\delta(q,a) , \boldsymbol{\sigma})
\end{array}
\end{equation}
where $a \in \Sigma$ is a symbol and $\boldsymbol{\sigma} \in \Sigma^*$ is a string, and $a\boldsymbol{x}$ is their concatenation.
$A$ accepts the string $\boldsymbol{\sigma}$ -i.e., $\boldsymbol{\sigma}$ is in the language of $A$, $L(A)$- if and only if $\delta^*(q_0, \boldsymbol{\sigma}) \in F$.
Let $\boldsymbol{\sigma} = (\sigma^{(1)}, \sigma^{(2)}, ..., \sigma^{(T)} )$ be the input string, we denote as $\boldsymbol{q} =(q^{(1)},q^{(2)}, ..., q^{(T)})$ the sequence of states visited by the automaton while processing the string, namely $q^{(0)} =q_0$ and $q^{(i)}=\delta(q^{(i-1)},\sigma^{(i)})$ for all $1\leq i \leq T$.

Moore machines are an extension of DFAs to non-binary outputs.
A Moore machine $M$ is a tuple $(\Sigma, Q, O,  q_0, \delta, \lambda)$, where $\Sigma$ is the alphabet of input symbols, $Q$ is the set of states, $O$ is a finite set of output symbols, $q_0 \in Q$ is the initial state, $\delta: Q \times \Sigma \to Q $ is the transition function, and $\lambda: Q \to O $ is the output function.

Given an input string $\boldsymbol{\sigma}$ the machine produces not only a sequence of states $\boldsymbol{q}$, but also an output string $\boldsymbol{o} = (o^{(1)},o^{(2)}, ..., o^{(T)})$, with $o^{(i)} \in O$ and $o^{(i)}=\lambda(q^{(i)})$ for all $1 \leq i \leq T$.

We can extend the output function over strings as we did for the transition function, which we denote as $\lambda^*$. We have that $\lambda^*:Q \times \Sigma^* \rightarrow O$ and
\begin{equation} \label{eq:rew_over_strings}
    \lambda^*(q, \boldsymbol{\sigma}) = \lambda(\delta^*(q,\boldsymbol{\sigma}))
\end{equation}
$\lambda^*(q, \boldsymbol{\sigma})$ is the \textit{last} output generated by the machine by processing the input sequence $\boldsymbol{\sigma}$ starting from state $q$.

\subsection{Probabilistic Finite Automata} \label{sec:PFA}
Probabilistic Finite Automata (PFA) generalize DFAs by making both the initial state and the transitions probabilistic, and uses a final (output) distribution over states to define acceptance.
A PFA $A_p$ is a tuple $(\Sigma, Q, \mu, \delta_{p}, \rho)$, where $\Sigma$ is the alphabet; $Q$ is the set of states, $\mu \in \Delta(Q)$ is the initial state distribution, i.e., \( \mu(q) = \Pr(q_0 = q) \); $\delta_{p}: Q \times \Sigma \times Q\rightarrow [0,1]$ is the transition probability function; and $\rho \in \Delta(Q)$ is the output (acceptance) probability function, i.e., \( \rho(q) = \Pr(\text{accept} \mid \text{end in } q)\).

The probabilistic transition function $\delta_p$ can also be represented more compactly in matrix form, as 3D \textit{transition matrix} $\mathcal{T} \in \mathbb{R}^{|\Sigma|\times|Q| \times |Q|}$, having $\mathcal{T}[\sigma, q, q'] = \delta_p(q, \sigma, q')$ $\forall$ $q,q' \in Q$ and $\sigma \in \Sigma$


This matrix representation is shown in Figure \ref{fig:PFAvsDFA}(b).

Given a string $\boldsymbol{\sigma}$, we denote as $ \boldsymbol{\tilde{q}}= \tilde{q}^{(1)}, \tilde{q}^{(2)},..., \tilde{q}^{(T)}$ the sequence over time of probabilities distributions on states. We have
\begin{equation}
\begin{array}{l}
    \tilde{q}^{(0)} = \mu \\
    \tilde{q}^{(i)} = \tilde{q}^{(i-1)} \mathcal{T}[\sigma^{(i)}]\quad\quad 1 \leq  i \leq T
\end{array}
\end{equation}
The probability of being in a final state at time $i$ is the inner product $\tilde{q}^{(i)} \cdot \rho^T$.

Since the probability of a string $\sigma$ to be accepted by the PFA, written as $Pr_{A_p}(\mathbf{\boldsymbol{\sigma}})$, is the probability to be in a final state in the last state $\tilde{q}^{(T)}$, it is calculated as follows
\begin{equation}
    Pr_{A_p}(\mathbf{\boldsymbol{\sigma}}) = \mu \cdot \mathcal{T}[\sigma^{(1)}]\cdot \mathcal{T}[\sigma^{(2)}]\cdot ... \cdot \mathcal{T}[\sigma^{(T)}] \cdot \rho^T\label{acceptance_formula}
\end{equation}
Figure \ref{fig:PFAvsDFA} shows a comparison between a PFA (b) and a DFA (c).

\subsection{Linear Temporal Logic over Finite Traces}
Linear Temporal Logic (LTL) \cite{LTL} is a language which extends traditional propositional logic with modal operators. With the latter we can specify rules that must hold \textit{through time}. In this work, we focus in particular on LTL interpreted over finite traces (LTLf) \cite{LTLf}. Such interpretation allows the executions of arbitrarily long traces, but not infinite, and is adequate for finite-horizon planning problems, especially for expressing goals and safety constraints.

Given a set $P$ of propositions, the syntax for constructing an LTLf formula $\phi$ is given by
\begin{equation}
    \phi ::= \top \mid \bot \mid p \mid \lnot \phi \mid \phi_1 \wedge \phi_2 \mid X\phi \mid \phi_1 \, U \, \phi_2
\end{equation}
where $p \in P$. We use $\top$ and $\bot$ to denote true and false respectively. $X$ (Next) and $U$ (Until) are temporal operators. Other temporal operators are: $N$ (Weak Next) and $R$ (Release) respectively, defined as $N \phi \equiv \lnot X\lnot \phi$ and $\phi_1 R \phi_2 \equiv \lnot(\lnot\phi_1 U \lnot \phi_2 )$; $G$ (globally) $G\phi \equiv \bot R\phi$ and $F$ (eventually) $F \phi \equiv T U \phi$.

We refer the reader to \cite{LTL} for a formal description of the operators' semantics.
Any LTLf formula $\phi$ can be translated into an equivalent Deterministic Finite Automaton (DFA) $A_\phi = (2^P, Q, q_0 , \delta, F)$.
which accepts all the traces satisfying the formula
\begin{equation}
    \boldsymbol{\sigma} \vDash \phi \text{ iff } \boldsymbol{\sigma} \in L(A_\phi)
\end{equation}
Although the size of $A_\phi$ is double-exponential in $\phi$ in the worst-case \cite{LTLf}, $A_\phi$ is often quite small in practice, and scalable techniques are available for computing it from $\phi$ \cite{LTL2DFA1, LTL2DFA2, LTL2DFA3}.


\subsection{Other Languages}
Our NeSy framework is capable of representing all regular grammars, also known as Type 3 grammars according to Chomsky’s hierarchy \cite{chomsky_dec}. These grammars can be equivalently described using deterministic finite automata (DFA) or regular expressions (RE).
It is worth noting that Linear Temporal Logic over finite traces (LTLf) can only express star-free regular languages, which form a proper subset of regular languages. An extension of LTLf, known as Linear Dynamic Logic over finite traces (LDLf), incorporates regular expressions as dynamic modalities for specifying temporal properties over finite traces. LDLf is expressive enough to capture the entire class of regular languages, while preserving desirable features such as a concise syntax and an effective translation into DFA.
Similar expressiveness results hold for the past-only variants of LTLf and LDLf, referred to as Past LTLf (PLTLf) and Past LDLf (PLDLf), respectively.
A summary of the expressive power of these logics is presented in Figure \ref{fig:PFAvsDFA}(a).

In our experiments, we primarily use DFA and LTLf to encode prior knowledge for integration into deep learning models. Therefore, we do not provide further details on the other formalisms here. Nonetheless, they are fully compatible with our framework. For formal definitions of LDLf and the past-only variants of LTLf and LDLf, we refer the reader to \cite{LTLf} and \cite{pLTLf}, respectively.
\begin{figure*}[t]
    \centering

\subfigure[]{
    \includegraphics[width=0.2\textwidth]{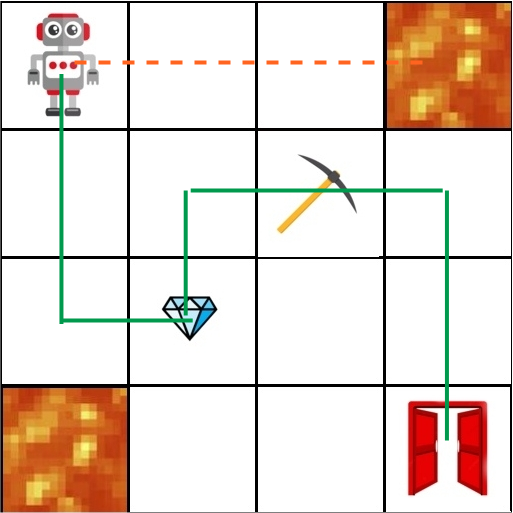}
    }
\subfigure[]{
    \includegraphics[width=0.23\textwidth]{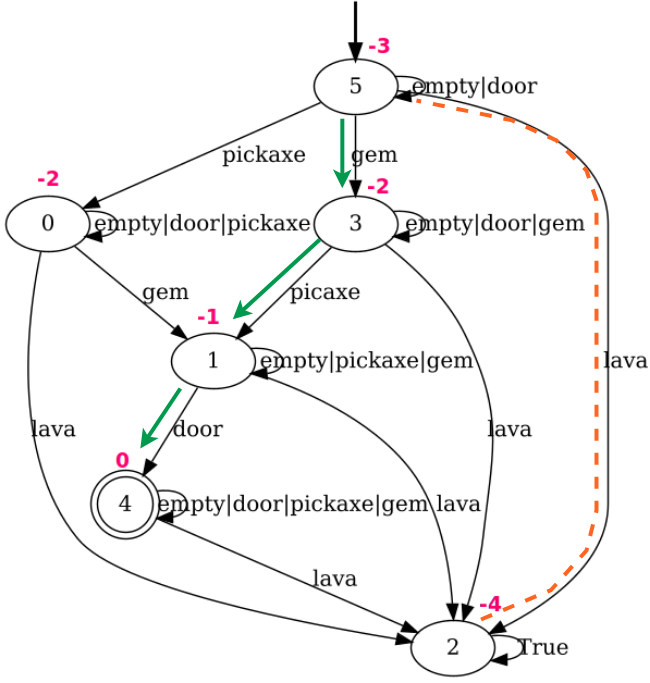}
    }

    \caption{Example of use of LTLf and DFA in non-Markovian RL. a) An example of non-Markovian navigation environment inspired by the Minecraft videogame. b) Moore Machine for the task: Moore Machine for the task: the agent has to visit the pickaxe and the gem (in any order), before reaching the door, while always avoiding the lava. We draw in green a trajectory accomplishing the task and in dashed red a trajectory failing the task, in both the environment (a) and the automaton (b). }
    \label{fig:examples}
\end{figure*}

\subsection{Reinforcement Learning of Temporally Extended Tasks}\label{sec:nrm_and_RL}

Temporal specifications are also popularly exploited in Reinforcement Learning (RL) to address non-Markovian environments. In standard RL \cite{sutton}, in fact, the agent-environment interaction is modeled as a Markov Decision Process (MDP).
An MDP is a tuple $(S,A,t,r,\gamma)$, where $S$ is the set (possibly infinite) of environment \textit{states}, $A$ is the set of agent's \textit{actions} (possibly continuous), $t: S \times A \rightarrow \Delta(S)$ is the \textit{transition function}, $r:S \times A \rightarrow \mathbb{R}$ is the \textit{reward function}, and $\gamma \in [0,1]$ is the \textit{discount factor} expressing the preference for immediate over future reward.

In this classical setting, transitions and rewards are assumed to be \textit{Markovian} – i.e., they are functions of the current state only.
Although this formulation is general enough to model most decision problems, it has been observed that many natural tasks are non-Markovian \cite{littman2017}.
A decision process can be non-Markovian because Markovianity does not hold for the reward function $r:(S\times A)^* \rightarrow \mathbb{R}$, or the transition function $t:(S \times A)^*\rightarrow \Delta(S)$, or both. In this work we focus mainly on non-Markovian \textit{Reward} Decision Processes (NMRDP) \cite{restrainingBolts}.
Learning an optimal policy in such settings is hard, since the current environment outcome depends on the entire history of state-action pairs the agent has explored from the beginning of the episode; therefore, regular RL algorithms are not applicable. Rather than developing new RL algorithms to tackle NMRDP, the research has focused mainly on how to construct Markovian state representations for the task at hand. An approach of this kind are the so called Reward Machines.

Reward Machines (RMs) are an automata-based representation of non-Markovian reward functions \cite{RM_journal}. Given a finite set of propositions $P$, representing abstract properties or events observable in the environment, RMs define temporally extended rewards over these propositions while exposing their compositional structure to the learning agent.
More specifically, a Reward Machine is a Moore machine $M = (\Sigma, Q, O, q_0, \delta, \lambda)$ used to assign reward values to agent trajectories. In particular, the input alphabet $\Sigma$ is either equal to $P$ or to $2^P$, depending on whether the propositions are mutually exclusive in the environment (i.e., whether the agent can observe multiple propositions as true at the same time).
The output set $O \subset \mathbb{R}$ consists of continuous values used to reward the agent upon reaching specific automaton states.
A key component of this framework is the \emph{labeling function} $L: S \rightarrow \Sigma$,
which maps environment states to input symbols for the Moore machine.
Let $\boldsymbol{s} = [s^{(1)}, s^{(2)}, \dots, s^{(T)}]$ be a sequence of states observed by the agent during its interaction with the environment. The labeling function is applied to each state to extract a sequence of symbols, or \emph{trace}:
\[
    \boldsymbol{\sigma} = (L(s^{(1)}), L(s^{(2)}), \dots, L(s^{(T)})).
\]
This trace is then processed by the Reward Machine to produce the current automaton state $q^{(T)}$ and the corresponding output reward $o^{(T)}$.
By leveraging the knowledge of both $M$ and $L$, we gain two important capabilities:
\begin{itemize}
    \item We can construct an \emph{augmented state representation} as the cross-product of environment and automaton states, $S \times Q$. This combined representation forms a Markovian state space for the task encoded by the Reward Machine \cite{restr_bolts}.
    
    \item We obtain an \emph{automated mechanism for defining reward functions} that guide the agent toward completing the task specified by the automaton. Specifically, we define the reward function $r : S^* \rightarrow \mathbb{R}$ as:
    \[
        r(\boldsymbol{s}) = \lambda^*(q_0, L(\boldsymbol{s})),
    \]
    where, with a slight abuse of notation, $L(\boldsymbol{s})$ denotes the entire sequence of symbols extracted from the state sequence $ L(\boldsymbol{s}) = (L(s^{(1)}), L(s^{(2)}), \dots, L(s^{(T)}))$.
\end{itemize}

\paragraph{Example: Image-Based Minecraft-Like Environment} \label{par:example}

Consider the environment shown in Figure \ref{fig:examples}(a), which consists of a grid world containing various items: a pickaxe, a gem, a door, and a lava cell. The robot in the grid must accomplish the following task:  visit at least once the pickaxe and the gem (in any order), before reaching the door, while always avoiding the lava. This task can be represented by the Moore machine depicted in Figure \ref{fig:examples}(b), defined over five symbols: one for each distinct item, plus a symbol indicating the absence of items. Thus, the input alphabet is $\Sigma = {\texttt{P}, \texttt{G}, \texttt{D}, \texttt{L}, \texttt{E}}$, where the letters stand for ‘pickaxe’, ‘gem’, ‘door’, ‘lava’ and ‘empty-cell’ in the order. Each symbol is considered true when the agent is currently located on the corresponding item cell, and false otherwise.
The agent must learn how to navigate the grid to satisfy the task. At each step, it receives a state $s \in S$ in the form of an image showing the agent's position in the grid, similar to the example shown in the figure. This state representation is not Markovian, since the current image only reveals which item (if any) the agent is on now, but not which items it has visited in the past. This non-Markovianity is resolved by tracking the machine state $q$. However, to do so, the agent must know the labeling function that maps image observations of the grid world to the input symbols in $\Sigma$.

\section{DeepDFA}
In this section, we present DeepDFA, our neurosymbolic framework for integrating temporal constraints representable as DFAs or Moore Machines. These automata can also originate from specifications written in LTLf or LDLf, which are then translated into DFAs, and possibly augmented with an output function so to become Moore Machines. This includes, for example, declarative specifications from business process models or non-Markovian RL tasks.

Our model is a hybrid between a recurrent neural network and an automaton. It can represent either a DFA or a Moore Machine, while remaining continuous and differentiable, making it seamlessly integrable into any neural network architecture for a wide range of sequential tasks. We present two versions of the model: a preliminary one designed for categorically grounded sequences, and an extended version capable of handling probabilistic or fuzzy input symbols. The latter is applicable to problems involving subsymbolic observations that require a perception phase—i.e., cases where the truth value of symbols must be inferred via a complex symbol grounding module.

The weights of our model can either be initialized to encode known DFAs, or learned from sequential data. In the former case, DeepDFA enables the injection of prior background knowledge; in the latter, it performs automata learning through neural training.
In this work, we focus on leveraging DeepDFA as a tool for knowledge injection. However, preliminary results on its use for automata learning are also promising \cite{deepDFA_ECAI24}, and we leave this direction for future work.

\subsection{Neural Architecture}
Here we describe in detail the NeSy architecture used for encoding automata knowledge as a neural model.
Our technique is grounded in the theory of Probabilistic Finite Automata (PFA) introduced in Section \ref{sec:PFA}. The main intuition behind our work is that PFAs are closely related to Recurrent Neural Networks (RNN), since they calculate the next state and output using multiplications between continuous vectors and matrices, in the same way RNNs do. However, DFAs can also be represented in matrix form, with the difference that their matrix representation is composed of only one-hot row vectors, as shown in the Figure \ref{fig:PFAvsDFA} (c).
Following this idea, we define DeepDFA as a parametric PFA, which can both be initialized with prior knowledge, or learned from experience.
Let $|Q|$, $|\Sigma|$, and $|O|$ be respectively the number of states, input, and output symbols of our automaton. 
This is the formulation of DeepDFA in case of categorically grounded input symbols.
\begin{equation} \label{eq:transition_rnn}
\begin{array}{l}
    \mu = \text{softmax}(\theta_{\mu}/ \tau) \\
    \mathcal{T} = \text{softmax}(\theta_{\mathcal{T}} / \tau) \\
    \mathcal{R} = \text{softmax}(\theta_{\mathcal{R}} / \tau) \\
    \\
    \tilde{q}^{(0)} = \mu \\
    \tilde{q}^{(i)} = \tilde{q}^{(i-1)} \cdot \mathcal{T}[\sigma^{(t)}]\\
    \tilde{o}^{(i)} = \tilde{q}^{i} \cdot \mathcal{R} \\

\end{array}
\end{equation}
$\tilde{q}^{(i)}$ and $\tilde{o}^{(i)}$ are the automaton state and output at time-step $i$, they are probability vectors of size $|Q|$ and $|O|$ respectively. $\sigma^{(i)} \in \Sigma$ is the current input symbol.
The model can be recursively applied on the symbols of an input trace $\boldsymbol{\sigma}= (\sigma^{(1)}, \sigma^{(2)}, ..., \sigma^{(T)})$, and produce $T$ belief vectors on the automaton states and outputs, one for each time step.
We denote with $\mu$, $\mathcal{T}$, and $\mathcal{R}$, respectively, the initial state distribution vector and the transition and the output matrices of the machine. Note that the notation follows the one given for PFAs in section \ref{sec:PFA}, except for the output distribution vector $\rho$, which is replaced here with the output \textit{matrix} $\mathcal{R \in \mathbb{R}^{|Q| \times |O|}}$. In this way the machine can predict more than two outputs. The semantic of the output distribution matrix follows the one of the output vector: $\mathcal{R}$ contains in position $[q,o]$ the probability of emitting the output $o$ while being in state $q$.  
The learnable model parameters are $\theta_{\mu} \in \mathbb{R}^{|Q|}$, $\theta_{\mathcal{T}} \in \mathbb{R}^{|\Sigma|\times|Q|\times |Q|}$ and $\theta_{\mathcal{R}} \in \mathbb{R}^{|Q|\times |O|}$.
Note that $\mu$, $\mathcal{T}$, and $\mathcal{R}$ are obtained from the neural network parameters through activation. 
The temperature $\tau$, $0 < \tau \leq 1$, is used to control the steepness of the activation functions. When $\tau$ is equal to one the model uses the normal activation functions, as the temperature tends to zero the softmax and sigmoid activations tend to the argmax and step function, respectively, and the model approximates more closely a DFA and less a PFA, with all the rows of the transition matrix and the output vector becoming nearly one-hot, as shown in Figure \ref{fig:PFAvsDFA} (c).  The activation is employed to ensure that $\mu$, $\mathcal{T}$ and $\mathcal{R}$ probability distributions \footnote{where no differently stated the activation is intended on the last dimension of tensors, in this case the softmax activation make each row of the matrices sum up to one.}, and the temperature is used to control and possibly limit stochasticity.
However they are both not needed in case of knowledge injection, since in this case the model parameters can be directly initialized to the matrix representation of the machine, which is known, and already one-hot. However, activation does not affect the model predictions in case of knowledge injection, therefore we let it in the neural network definition, to standardize the notation to the learning case, where the machine is learned instead of injected.

In particular, knowledge injection of a Moore Machine $(\Sigma, Q, O,  q_0, \delta, \lambda)$ is achieved by initializing the network parameters $\theta_{\mu}$, $\theta_{\mathcal{T}}$ and $\theta_{\mathcal{R}}$ in the following way.
\begin{equation}
\begin{aligned}
    \theta_\mu[i] &= 
    \begin{cases}
        1 & \text{if } q_i = q_0 \\
        0 & \text{otherwise}
    \end{cases} \\
    \theta_{\mathcal{T}}[i, j, k] &=
    \begin{cases}
        1 & \text{if } \delta(q_j, s_i) = q_k \\
        0 & \text{otherwise}
    \end{cases} \\
    \theta_{\mathcal{R}}[i, j] &= 
    \begin{cases}
        1 & \text{if } \lambda(q_i) = o_j \\
        0 & \text{otherwise}.
    \end{cases}
\end{aligned}
\end{equation}

DeepDFA  allows us to represent uncertainty over both the transition and the output function, because it is based on representing the model as a probabilistic machine.
However, note that the input symbols $\sigma$ are \textit{categorically grounded}. In particular the current symbol $\sigma^{(i)}$ is used to index the transition matrix in Equations \ref{eq:transition_rnn} and, as a consequence, must be represented as an integer. This contrasts with neural network-based symbol grounding techniques, which usually predict a \textit{probabilistic belief} on the symbol truth values.
For this reason, we extend the framework to be fully probabilistic, by modifying the current state computation as follows.

\begin{equation} \label{eq:deepDFA_uncertain_sym}
\begin{array}{l}
    \tilde{q}^{(i)}= \sum\limits_{j=1}^{j=|\Sigma|} \tilde{\sigma}^{(i)}_j (\tilde{q}^{(i-1)} \cdot \mathcal{T}[j] )\\
\end{array}
\end{equation}
Here the current symbol $\tilde{\sigma}^{(i)}$ is probabilistically grounded -note the tilde- therefore it consists in a probability vector of size $|\Sigma|$.
In order to calculate the state after receiving the current symbol we compute \textit{all the possible states} where we could end up with one of the symbols in $\Sigma$, and we weight them by the probabilities contained in the grounding $\tilde{\sigma}^{(i)}$. This weighted sum becomes the new belief vector on the automaton state.

The final model can be seen as two parametric functions. One takes the sequence of beliefs over the symbols alphabet $\boldsymbol{\tilde{\sigma}}$ and returns the sequences of machine states $\boldsymbol{\tilde{q}}$, and the other takes as input $\boldsymbol{\tilde{\sigma}}$ and produces the sequence of machine outputs $\boldsymbol{\tilde{o}}$. We denote them respectively as $\mathbf{Q}: \Delta(\Sigma)^* \rightarrow \Delta(Q)^*$, and $\mathbf{{O}}: \Delta(\Sigma)^* \rightarrow \Delta(O)^*$. We can write our model lifted on sequences as

\begin{equation}
    \begin{array}{l}
       \boldsymbol{\tilde{q}} = \mathbf{Q}(\boldsymbol{\tilde{\sigma}}| \theta_\mu, \theta_{\mathcal{T}})   \\
       \boldsymbol{\tilde{o}} = \mathbf{O}(\boldsymbol{\tilde{\sigma}} | \theta_{\mu}, \theta_{\mathcal{T}}, \theta_{\mathcal{R}})
    \end{array}
\end{equation}

\subsection{Relation to Other Frameworks in the Literature}

Here we describe the main differences between our framework, DeepDFA, and related approaches in the literature, in particular FuzzyDFA~\cite{Umili2023KR} and NeSyA~\cite{NeSyA}.

All three approaches are based on representing temporal knowledge in the form of automata. DeepDFA and NeSyA adopt a probabilistic semantics, whereas FuzzyDFA relies on a fuzzy semantics.

Another important difference concerns the type of automata employed. FuzzyDFA and NeSyA use \textit{symbolic} automata, while DeepDFA is based on \textit{simple} automata. The distinction lies in how traces are generated from the domain.

Let $P$ be a finite set of observable symbols in the domain (e.g., $P = \{a, b\}$). Simple automata are designed to monitor simple traces, i.e., strings such as $(a, b, b)$, where at each time step exactly one symbol is true and all others are false. This assumption—also known as the \emph{simplicity assumption}~\cite{chiariello}—naturally holds in many domains, such as natural language processing (NLP) and business process management.
Symbolic automata, on the other hand, are designed to monitor traces in which, at each time step, any number of symbols may be true (e.g., $(\{a\}, \{a,b\}, \{\})$). Such automata can always be translated into simple automata with alphabet $\Sigma = 2^P$.
The choice between simple and symbolic automata is therefore strongly domain-dependent. When the domain satisfies the simplicity assumption, representing knowledge using simple automata is preferable, since explicitly encoding this assumption within the knowledge representation would unnecessarily complicate the automaton.\footnote{In particular, when the automaton is obtained by translating an \textsc{LTL}\textsubscript{f} formula, exploiting the simplicity assumption has been shown to yield more efficient translations~\cite{chiariello}.}
Conversely, when symbols are not mutually exclusive, representing the domain with simple automata generally requires an exponential number of symbols with respect to the number of symbols in a symbolic automaton. In such cases, symbolic automata are typically preferable. However, in practice, symbolic variables are often not fully independent, and these dependencies can be exploited to construct a significantly smaller set of symbols than $2^P$, as we will show in a real-world case study in our experiments (see Section~\ref{sec:caviar_dataset}).

A further substantial difference between DeepDFA and both NeSyA and FuzzyDFA is that DeepDFA is the only framework that models probabilities in both the transition and the output functions of the automaton. As a consequence, although we evaluate it here in deterministic settings, DeepDFA can, in principle, be used to impose probabilistic knowledge, and in particular any knowledge that can be expressed as a probabilistic finite automaton (PFA).

Finally, our model is the only one in the literature that is fully learnable. Although we do not focus on this aspect in the present paper, extensive experimental evidence shows that DeepDFA can be learned directly from data~\cite{deepDFA_ECAI24}. This property opens the door to a wide range of additional reasoning–learning scenarios beyond fixed knowledge injection, which we leave for future work and which are not achievable with existing approaches that do not support learning the knowledge itself.

For completeness, Table~\ref{tab:comparison} summarizes all the differences among the considered frameworks.

\begin{table}[t]
\centering
\caption{Comparison between DeepDFA and related frameworks.}
\label{tab:comparison}
\begin{tabular}{lcccc}
\toprule
\textbf{Method} & \textbf{Semantics} & \textbf{Automaton} & \textbf{Stochastic Knowledge Support} & \textbf{Learnable} \\
\midrule
DeepDFA  & Probabilistic & Simple    & $\checkmark$ & $\checkmark$ \\
NeSyA    & Probabilistic & Symbolic  & $\times$  &  $\times$ \\
FuzzyDFA & Fuzzy         & Symbolic  & $\times$  & $\times$  \\
\bottomrule
\end{tabular}
\end{table}

\section{Learning and Reasoning with DeepDFA}
With the extension in Equation \ref{eq:deepDFA_uncertain_sym}, we can compute both the probability of being in a specific machine state $q$ and the probability of producing an output $o$ at a given point in the execution of the machine, by feeding the latter with probabilistically grounded sequences. This removes the unrealistic assumption—common in many logic-based systems—that symbols are perfectly grounded.
Moreover, this probability is differentiable, and can thus be used in the optimization of neural networks via gradient descent.

In this paper, we evaluate the ability of DeepDFA to be integrated into a typical NeSy pipeline encompassing both perception and reasoning. Specifically, we test DeepDFA on a Semi-Supervised Symbol Grounding (SSSG) task \cite{DeepProbLog, LTN, NeSyA}. The task consists in exploiting prior knowledge expressed as a logical formula to infer a perception function mapping raw observations (e.g., images, sensor readings, etc) to input symbols.
Of course, this is not the only possible application of DeepDFA. For example, we are also working on its integration into tasks involving autoregressive symbol generation \cite{umili_ltlf_bpm}. However, SSSG is a widely adopted benchmark in the NeSy literature, and we focus on it in this work. Furthermore, we show that semi-supervised symbol inference can serve as a fundamental building block for several machine learning applications. Two such applications are (i) static classification of subsymbolic sequential data, and (ii) learning to execute temporally extended tasks in subsymbolic environments via interaction and reinforcement learning.

The remainder of this section is organized as follows: (i) we first provide a formal definition of the SSSG problem, and we introduce the first application of interest: image stream classification ; (ii) we conclude by describing our how SSSG can be integrated into non-Markovian RL applications.
\begin{figure}[t]
    \centering
    \includegraphics[width=0.5\linewidth]{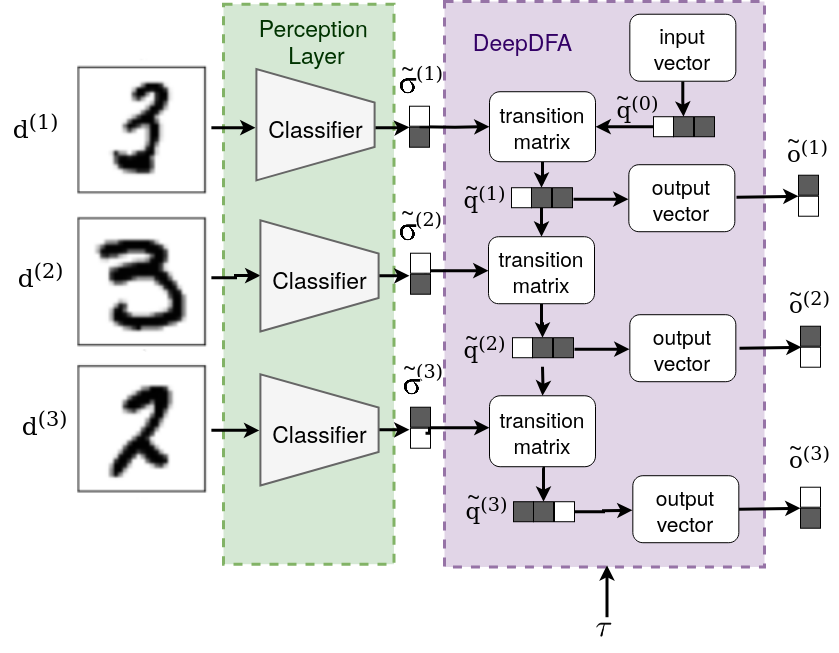}
    \caption{Neural architecture used for semi-supervised symbol grounding through knowledge injection with DeepDFA}
    \label{fig:sssg}
\end{figure}

\subsection{Semisupervised Symbol Grounding}
To employ DeepDFA for semi-supervised symbol grounding, we assume the existence of a sequential and subsymbolic process that is internally governed by a regular language or a Moore Machine.

This process takes as input a sequence of subsymbolic observations $\boldsymbol{d} = (d^{(1)}, d^{(2)}, \dots, d^{(T)})$, where each $d^{(i)} \in \mathcal{D}$ and $\mathcal{D}$ denotes the domain of raw observations, which may be continuous and unstructured. The process produces symbolic or categorical outputs $y$.

We model this process as a Neuro-Symbolic (NeSy) function, i.e., a composition of two components: a perception or grounding phase, and a symbolic reasoning procedure. The symbolic component is assumed to be known a priori, whereas the perception function is unknown and must be inferred by leveraging both the system's output $y$ and the available symbolic knowledge.

A practical application illustrating this process will be provided in the next section, to make the approach more concrete.

\subsection{Application 1: Image Stream Classification}

We consider the problem of classifying a sequence of images $\boldsymbol{d}$ as either belonging or not belonging to a regular language $L(A)$, where $A = (\Sigma, Q, \delta, q_0, F)$ is a deterministic finite automaton (DFA).
We assume that each image in the sequence is a visual rendering of a character from the automaton's alphabet $\Sigma$. In other words, there exists a target, exact symbol grounding function $g : \mathcal{D} \rightarrow \Sigma$, which categorically maps each image $d^{(i)}$ to an alphabet symbol $\sigma^{(i)} = g(d^{(i)})$.
By applying this grounding function to a stream of images, we obtain a symbolic trace $\boldsymbol{\sigma} = (\sigma^{(1)}, \sigma^{(2)}, \dots, \sigma^{(T)})$, which can then be provided as input to the automaton. The automaton outputs a classification label indicating whether or not the trace belongs to the language $L(A)$. Thus, the overall process can be viewed as a composition of the grounding function $g$ and the automaton $A$.

We assume that the entire process is not fully known, and our goal is to learn an approximation of the grounding function $g$ by leveraging knowledge of the automaton $A$.
To this end, we also assume access to a dataset of pairs $(\boldsymbol{d}, y)$ sampled from the target process, where $y$ is the ground-truth class label (positive if and only if $\boldsymbol{\sigma} \in L(A)$).

We can address this problem using DeepDFA by constructing a two-layer architecture. The first layer is a neural classifier $g(d \mid \theta_{g})$ with trainable parameters $\theta_{g}$, responsible for grounding individual images. The second layer is a DeepDFA model that processes uncertain input symbols, as described in Equation \ref{eq:deepDFA_uncertain_sym}.

We define the grounding function over sequences as $\boldsymbol{G}: \mathcal{D}^* \rightarrow \Delta(\Sigma)^*$, where:
\begin{equation}
\boldsymbol{G}(\boldsymbol{d}) = (g(d^{(1)}), g(d^{(2)}), \dots, g(d^{(T)}))
\end{equation}

The full architecture, illustrated in Figure~\ref{fig:sssg}, is formalized as follows:
\begin{equation} \label{eq:sssg_model}
\begin{array}{l}
\boldsymbol{\tilde{\sigma}} = \boldsymbol{G}(\boldsymbol{d} \mid \theta_g) \\
\boldsymbol{\tilde{o}} = \mathbf{O}(\boldsymbol{\tilde{\sigma}} \mid \theta_{\mu}, \theta_{\mathcal{T}}, \theta_{\mathcal{R}})
\end{array}
\end{equation}

We train the model by supervising the last output of the automaton sequence, $\tilde{o}^{(T)}$, using a standard classification loss—such as cross-entropy.
\begin{equation}
    \mathcal{L} = \text{cross-entropy}(\tilde{o}^{(T)}, y)
\end{equation}

\subsection{Application 2: Non-Markovian Reinforcement Learning} \label{sec:integration_with_RL}

\begin{figure*}[t]
    \centering

\subfigure[]{
    \includegraphics[width=0.95\textwidth]{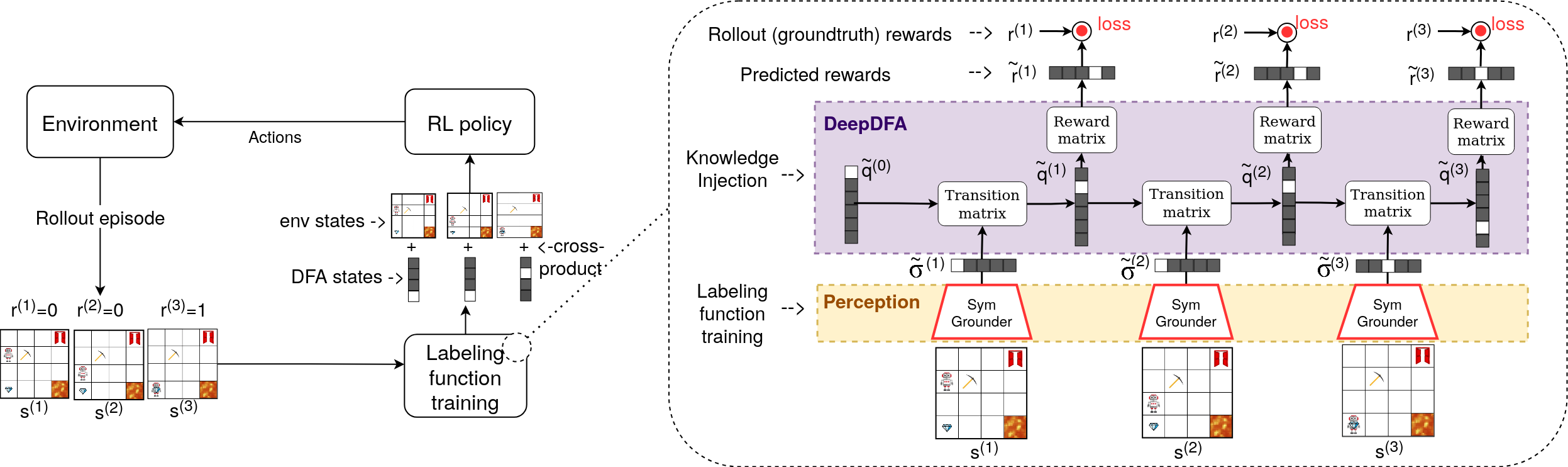}
}
    \caption{A visual representation of our algorithm for integrating knowledge through DeepDFA within non-Markovian RL}
    \label{fig:RL_application}
\end{figure*}
In Section \ref{sec:nrm_and_RL}, we discussed a classical system based on Reward Machines for learning temporally extended tasks.
In this application, we represent the task using a DeepDFA and jointly learn both the perception function—also referred to as the labeling function in the Reward Machine literature—and the policy required to solve the task.

In particular, we assume that an agent has to learn to perform a temporally extended task by optimizing a non-Markovian reward signal, which can be encoded as a Moore Machine. As is standard in RL, the agent interacts with the environment by taking an action and it receives a new state of the environment and a new reward in response. Additionally, we assume that the agent has \textit{purely symbolic} knowledge of the task it is executing. For example, in the Minecraft example depicted in Figure \ref{fig:examples}, and described in section \ref{sec:nrm_and_RL}, the agent would have access to the Moore Machine shown in Figure \ref{fig:examples}(b), but it would lack knowledge on \text{how to recognize} the pickaxe, the gem, and the other items within the environment. In this setting, traditional RMs are inapplicable because of the missing symbol grounding function, while purely deep-learning-based approaches have no way to exploit the symbolic knowledge and will rely only on rewards.
With our framework we can exploit both symbolic knowledge and data from the environment by interleaving RL and semi-supervised symbol grounding of the NRM. We now describe how we do that step by step.

First, we define the two-layer architecture used for semi-supervised grounding, introduced in Equation \ref{eq:sssg_model}. We initialize the DeepDFA parameters using the matrix representation of the prior symbolic knowledge available for the task, and we keep these parameters frozen during training. In contrast, the parameters of the symbol grounder, $\theta_{sg}$, are randomly initialized.

In this specific application, the input data sequence correspond to the sequence of environment states $\boldsymbol{s} = (s^{(1)}, s^{(2)}, \dots, s^{(T)})$ collected during an episode. As supervision for the model outputs, we use the sequence of rewards obtained in the environment, $\boldsymbol{r} = (r^{(1)}, r^{(2)}, \dots, r^{(T)})$.
Notably, this setup provides \textit{dense supervision}, as model predictions are supervised \textit{at each time step}.

During agent–environment interaction, we record each episode as a pair of sequences $(\boldsymbol{s}, \boldsymbol{r})$.
At regular intervals, we update the symbol grounder by training on this growing dataset of experience, minimizing the cross-entropy loss between model predictions and ground-truth rewards observed in the environment (learning step):
\begin{equation}
\mathcal{L} = \text{cross-entropy}(\boldsymbol{\tilde{o}}, \boldsymbol{r})
\end{equation}
Compared to the application of SSSG for static sequence classification, the dataset here is dynamic—it is constructed online. As a result, the quality of the data is highly dependent on the quality of the policy being executed. Since the policy starts off as random and improves over time through experience, early data collected during exploration is likely to be of low quality.

Another challenge we face is the non-Markovian nature of the task. To address this, following the principles of Reward Machines, we train the reinforcement learning agent using a state representation that combines the current environment state $s$ with the current automaton state $q$.
More specifically, at each step $i$ in the sequence, we construct the input to the RL policy by concatenating features extracted from $s^{(i)}$—which are learned online—with the probabilistic belief over the machine state $\tilde{q}^{(i)}$.
We compute belief vectors over automaton states using DeepDFA as follows (reasoning step):
\begin{equation}
\begin{array}{l}
\boldsymbol{\tilde{\sigma}} = \boldsymbol{G}(\boldsymbol{s} \mid \theta_g) \
\boldsymbol{\tilde{q}} = \mathbf{Q}(\boldsymbol{\tilde{\sigma}} \mid \theta_\mu, \theta_{\mathcal{T}})
\end{array}
\end{equation}

Since the symbol grounder is initially randomized, the predicted machine state initially deviates from the ground truth, and the state representation is not perfect. However,
as the agent observes more scenarios in the environment, the symbol grounding function of the model gradually becomes more similar to the unknown target grounding, and so does the distribution of machine states. In case the target labeling function is learned, our approach becomes equivalent to a RM.

A diagram of the procedure used for integrating DeepDFA within non-Markovian RL is shown in Figure \ref{fig:RL_application}
\subsubsection{On-Policy versus Off-Policy RL}
The training scheme outlined in the previous section constitutes our core algorithm for integrating DeepDFA in RL. While it can be used as-is with all on-policy algorithms, integrating it with off-policy algorithms requires some additional considerations.
Specifically, off-policy algorithms rely on a replay buffer to store transitions, which are then randomly sampled to compute updates for the reinforcement learning models. In the case of traditional Reward Machines with known grounding, transitions are stored in the replay buffer in the form $\boldsymbol{((}s^{(i)}, q^{(i)}\boldsymbol{)}, a^{(i)}, r^{(i)}, \boldsymbol{(}s^{(i+1)}, q^{(i+1)}\boldsymbol{)}, r^{(i+1)}\boldsymbol{)}$. If the same mechanism were applied in our setting — namely, by storing transitions of the form $\boldsymbol{((}s^{(i)}, \tilde{q}^{(i)}\boldsymbol{)}, a^{(i)}, r^{(i)}, \boldsymbol{(}s^{(i+1)}, \tilde{q}^{(i+1)}\boldsymbol{)}, r^{(i+1)}\boldsymbol{)} $- the automaton states \(\tilde{q}^{(i)}\) and \(\tilde{q}^{(i+1)}\) would be computed using the grounding function available at the time the transition is inserted into the replay buffer. Crucially, this grounding function may differ from the one used at the time the transition is later retrieved for training.
This issue does not arise in on-policy algorithms, where data are generated and immediately consumed for learning, ensuring that the reasoning step is always performed using the most up-to-date grounding function. Similar considerations apply to counterfactual experience generation, and in particular to Counterfactual Reward Machines (CRM) \cite{RM_journal}, a model-based technique that augments the replay buffer with transitions that are not actually experienced by the agent but are instead synthetically generated using an environment model. Unless the grounding function is guaranteed to be correct, using it to generate additional synthetic experience may adversely affect the learning process.

In summary, although integrating our model with off-policy or model-based reinforcement learning algorithms is not inherently infeasible, it requires additional care and methodological refinements. We therefore leave this integration as an interesting direction for future work.

\begin{figure*}[t]
    \centering

\subfigure[]{
    \includegraphics[width=0.2\textwidth]{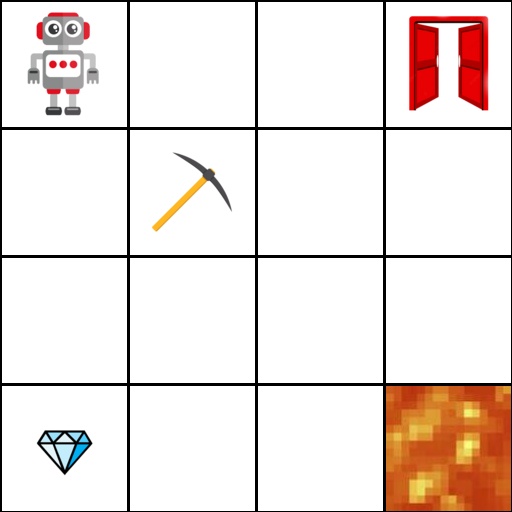}
    }
    \subfigure[]{
    \includegraphics[width=0.21\textwidth]{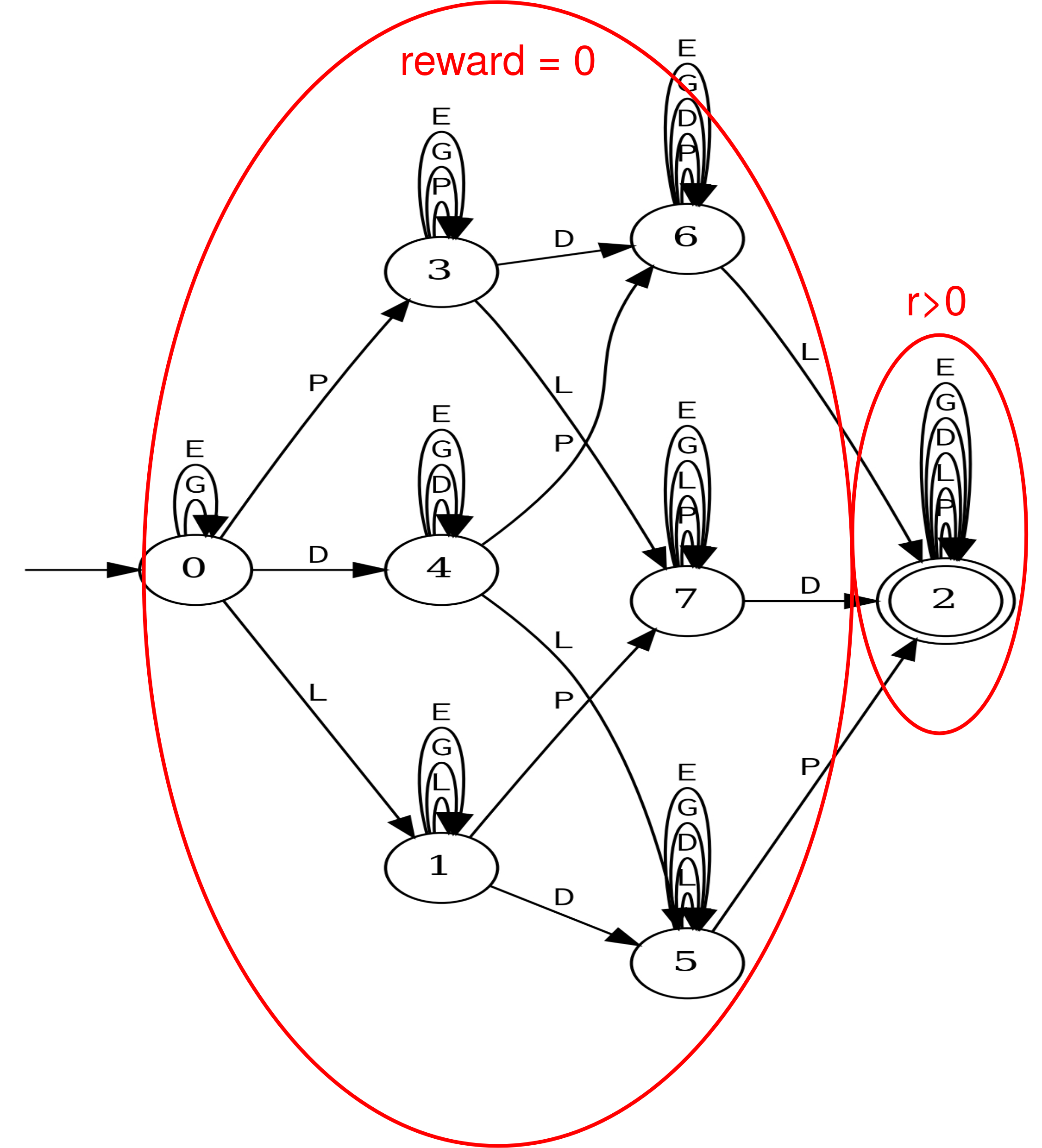}
    }
\subfigure[]{
    \includegraphics[width=0.23\textwidth]{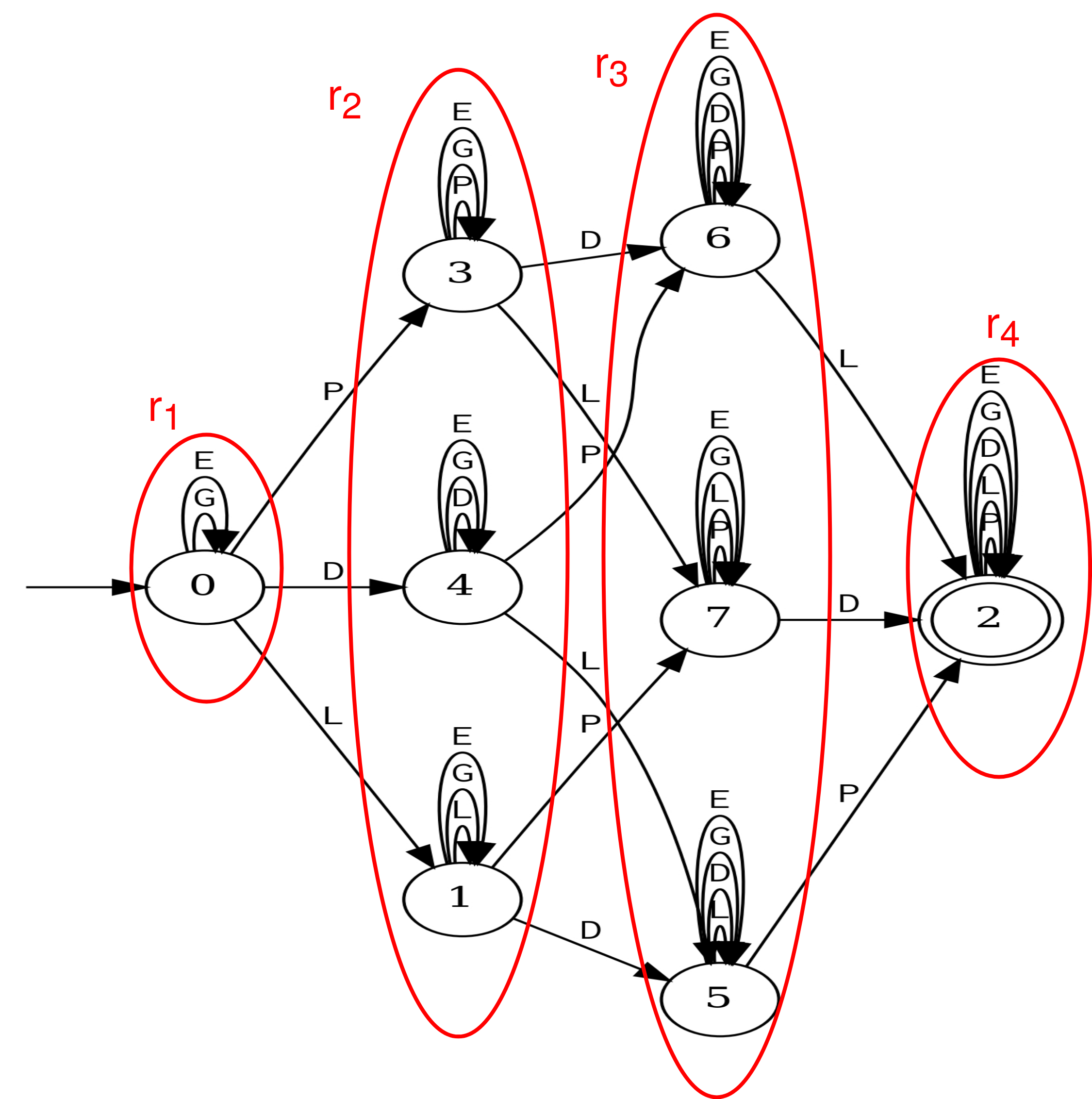}
    }

    \caption{Example of different automata-based reward schemes. a) Environment for LTLf navigation tasks. b -c) Moore Machine for the task: Moore Machine for the task: the agent has to visit the pickaxe (P), the lava (L) and the door (D) cells in any order. In (b) the machine is labeled with a sparse reward signal that only distinguish the final state from the others. In (c) the machine is labeled using a dense reward signal based on the distance from the final state}
    \label{fig:rew_as_supervision}
\end{figure*}
\subsubsection{Reward Values as Supervision} \label{sec:rew_as_supervision}
Since the reward values serve as supervision for DeepDFA, we have observed that overly sparse rewards are not very effective for symbol grounding. For instance, consider the automaton of the minecraft grid world example, shown in Figure \ref{fig:rew_as_supervision}(b). The automaton has 8 states, of which only one is final (state 2). If we employ a sparse signal, shown in Figure \ref{fig:rew_as_supervision}, rewarding the agent solely upon task completion, this entails assigning a positive reward to state 2 and 0 reward to all the other states. Subsequently, all the unsuccessful episodes would provide no feedback to the symbol grounder. Moreover, in case the agent accomplishes the task, the grounder would glean feedback solely to discern that the \textit{last} observation leading to winning reward ought to be associated with one of the three symbols bringing from a non-rewarding state to state 2: lava (L), door (D), or pickaxe (P). However, it would lack the means to distinguish among these three symbols in the last observation, or discern which symbols the agent has encountered in the previous observations. If instead we assign rewards to the states based on the distance from the nearest final state, Figure \ref{fig:rew_as_supervision}(c), this partitions the machine states in figure according to 4 reward values, $r_1, r_2, r_3$ and $r_4$. This kind of potential-based reward shaping is very commonly applied in RM applications to facilitate policy learning through RL \cite{RM_journal}. Note that the reward shaping generally does not provide distinct feedback for each single state of the machine. Consequently, while the resulting grounding process becomes more feasible, it is by no means straightforward in any case.

Note that this is a fundamental problem in neurosymbolic AI, which does affect all the estimators semisupervised by logical knowledge \cite{resoning_shrotcuts_survey}, not only our framework. For this reason we will explore in our experiments both sparse and dense reward schemes, in order to investigate how they affect RL performances. 

\section{Evaluation}
In this section, we report the experiments validating our framework.
We conduct experiments on using DeepDFA for injecting knowledge in two different application: image-stream classification and non-Markovian RL, that were previously described in detail in section. Our code is available at \url{https://github.com/KRLGroup/DeepDFAGrounding} and \url{https://github.com/KRLGroup/NeuralRewardMachines}\footnote{The code is currently split across two repositories, but we will consolidate it into a single codebase upon publication of the paper.}
\subsection{Image Stream Classification}
\label{sec:imagestream}

First of all, we test the capabilities of DeepDFA to ground symbols for a static image sequence classification problem. The code used to produce these results is available at \url{https://github.com/KRLGroup/DeepDFAGrounding}

\subsubsection{Synthetic image stream tagging: MNIST Digits for Declare}
A popular benchmark for semisupervised symbol grounding is the digit addition problem \cite{DeepProbLog}, where a system must learn to classify MNIST digits images by knowing only the result of their sum and how addition works.
This benchmark is not appropriate for our approach, since the addition task is not extended in time.
For this reason, we propose a similar benchmark in \cite{Umili2023KR} based on MNIST digits and LTLf formulas as logical knowledge.

In particular, we selected the Declare language as target formulas, which consists in a set of 21 LTLf formulas expressing constraints of various types that are commonly used in BPM. For the complete list of Declare formulas, we refer the reader to \cite{Declare_cons}. Each formula is translated into an equivalent DFA using the LTL2DFA translation tool \footnote{\url{https://github.com/whitemech/LTLf2DFA}}.
Since Declare formulas are defined over the binary alphabet $P=\{a,b\}$, we construct the datasets by representing symbolic configurations using images of zeros and ones from the MNIST dataset \cite{MNIST_dataset}. For each formula, we select all the traces of length between 1 and 5 as training traces, and we construct two test sets with traces having length respectively equal to 10 and 15, in order to assess the capability of our framework to generalize to longer sequences. We label each trace as accepted or rejected. Note that we supervise only the output in the \textit{last step} of the sequence. After that, we balance the test sets between positive and negative samples, removing excessive labels, while we let the training set imbalanced. We construct the final image datasets by replacing each symbol in the training (test) traces with an image from the MNIST training (test) set.
It is important to note that the test set includes symbolic patterns that were not seen during training, with each symbol rendered as an image never encountered in the training set.

\begin{figure*}
    \subfigure[]{\includegraphics[width=0.35\textwidth]{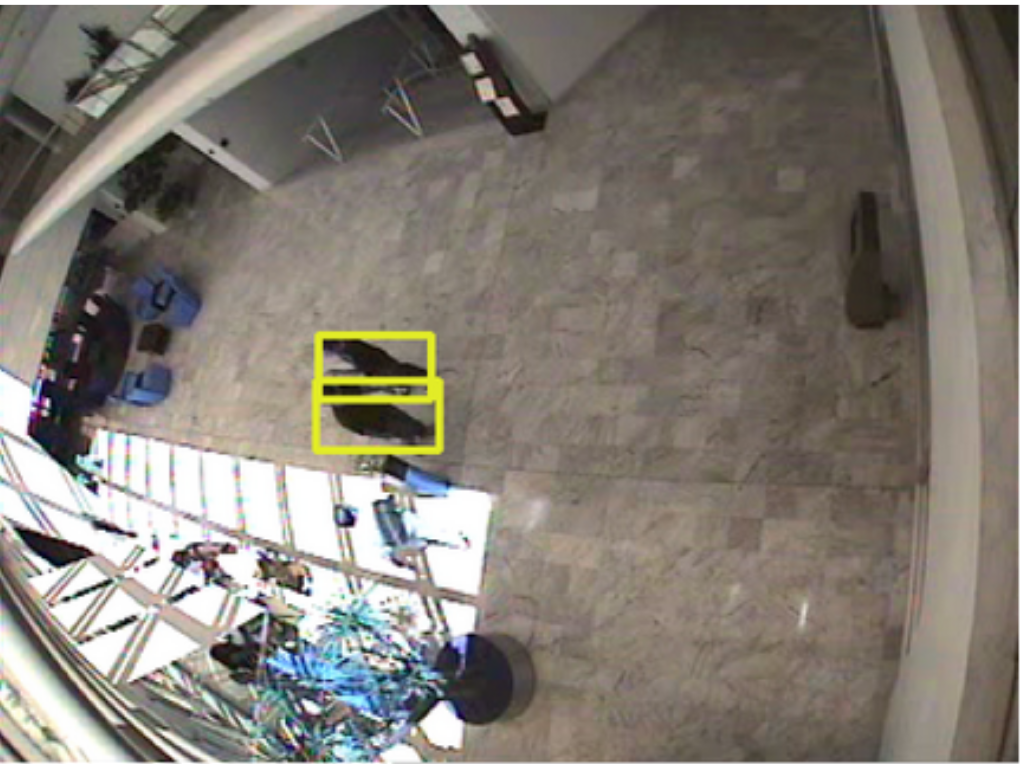}
    }
    \subfigure[]{\includegraphics[width=0.4\textwidth]{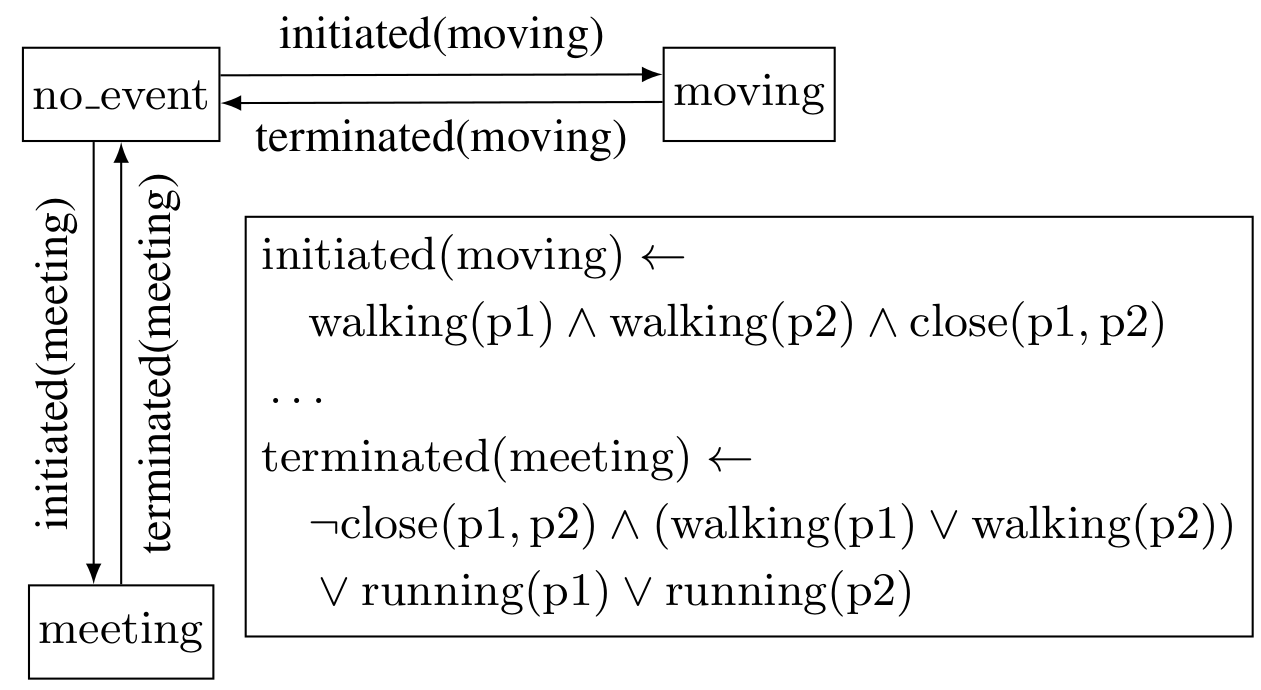}
    }
    \subfigure[]{
    \includegraphics[width=0.78\textwidth]{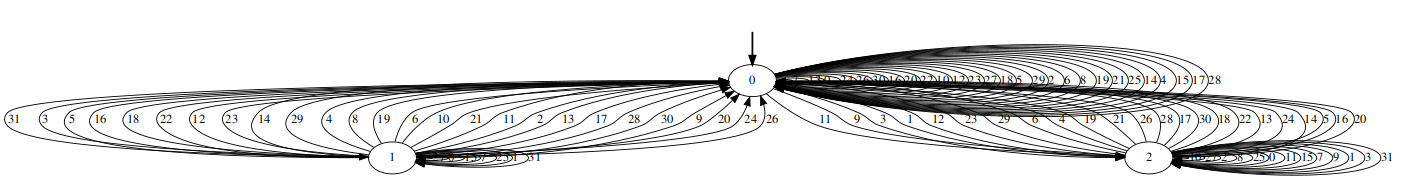}
    \includegraphics[width=0.21\textwidth]{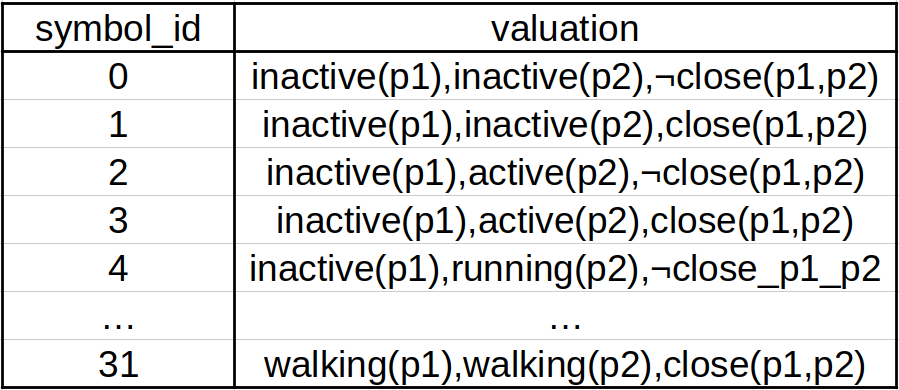}
    }
\caption{a) A frame from the CAVIAR dataset. b) Prior logical knowledge on the event recognition task in CAVIAR, image taken from \cite{NeSyA}. c) Simple DFA obtained from (a) with 32 symbols, and semantic for each newly introduced symbol. }\label{fig:caviar}
\end{figure*}

\subsubsection{Event Recognition in the CAVIAR dataset} \label{sec:caviar_dataset}

We also evaluate DeepDFA on a \emph{complex event recognition} task using public-space surveillance videos. Specifically, we use the CAVIAR benchmark dataset, which consists of surveillance videos of a public space and is publicly available at \url{https://homepages.inf.ed.ac.uk/rbf/CAVIARDATA1/}. The videos are staged and depict actors walking around, sitting down, meeting one another, leaving objects behind, and performing other activities. A frame example is shown in Figure \ref{fig:caviar}(a). Each video has been manually annotated with bounding boxes; \emph{short-term activities}, i.e., activities occurring over a short time span and detected at the level of individual video frames; and \emph{long-term activities}, which are temporally extended and can only be identified from multiple consecutive frames.
Short-term activities include events such as a person or object entering or exiting the surveillance area, walking, running, moving abruptly, or being active or inactive. Examples of long-term activities include a person leaving an object unattended, two people meeting, moving together, or fighting.
In this experiment, we investigate whether injecting prior knowledge about how long-term activities are determined by specific temporal patterns of short-term activities can improve the classification of long-term activities.
We follow the experimental setting of \cite{NeSyA}. In particular, we focus on the classification of three long-term events - \emph{moving}, \emph{meeting}, and \emph{no event} - which are the most frequently occurring in the dataset. We consider only videos in which two actors are present, denoted as $p_1$ and $p_2$. For each sequence, we exploit the bounding boxes of the people in the scene provided by the dataset. The final dataset consists of 8 training sequences and 3 test sequences. The label distribution in the training set comprises 1,183 frames of \emph{no event}, 851 frames of \emph{moving}, and 641 frames of \emph{meeting}. In the test set, the corresponding counts are 692, 256, and 894 frames, respectively. The mean sequence length is 411 frames, with a minimum length of 82 and a maximum length of 1,054.

The prior temporal knowledge used for this task is encoded as a three-state automaton, capturing a variant of the Event Calculus \cite{event_calculus} programs for CAVIAR presented in \cite{event_calculus_for_event_recognition}, and is shown in Figure~\ref{fig:caviar}(b). Short-term activities are treated as input symbols of the automaton, while long-term activities label the automaton states. The three long-term events are described in terms of nine short-term activities:
$inactive(p)$, $active(p)$, $walking(p)$, and $running(p)$, with $p \in \{p_1, p_2\}$ (yielding eight symbols), plus the predicate $close(p_1, p_2)$, which denotes whether the two actors are spatially close.
The CAVIAR specification is given as a \emph{symbolic nondeterministic automaton}. Symbolic automata have transitions labeled with \emph{logical formulas} (e.g., $a \wedge b$), whereas the simple automata considered in this paper (see Section~\ref{sec:DFA}) have transitions labeled with atomic propositions (e.g., $a$, $b$). Given a set $P$ of propositional symbols observable at each time step, symbolic automata have alphabet $\Sigma = 2^{P}$, while simple automata have alphabet $\Sigma = P$. Any symbolic automaton can be translated into a simple one by mapping each possible interpretation over $P$ to a distinct symbol.
We converted the CAVIAR automaton into a \emph{simple deterministic automaton} by exploiting the observation that, although $|P| = 9$ would in principle yield $2^{9} = 512$ possible symbols, not all interpretations of $P$ are semantically feasible. For instance, a person cannot simultaneously walk and run. Given the semantics of the short-term activity predicates, only 32 interpretations are possible, corresponding to
$4$ activity classes for $p_1$ $\times$ $4$ activity classes for $p_2$ $\times$ $2$ possible values of $close(p_1, p_2)$. Moreover, when restricted to these 32 interpretations, the automaton is deterministic. This allows our framework to be applied without modification. Figure~\ref{fig:caviar}(c) shows the resulting DFA and the semantics of each newly introduced symbol.

\begin{figure*}[t]
    \centering
    \subfigure[]{
    \includegraphics[width=0.32\textwidth]{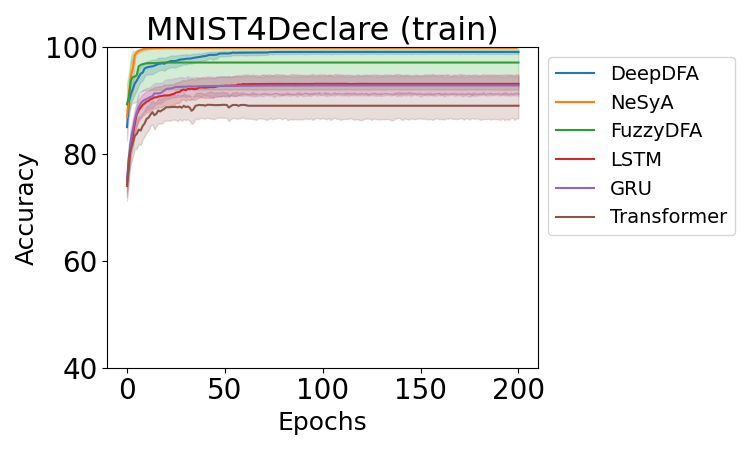}}
\subfigure[]{
    \includegraphics[width=0.32\textwidth]{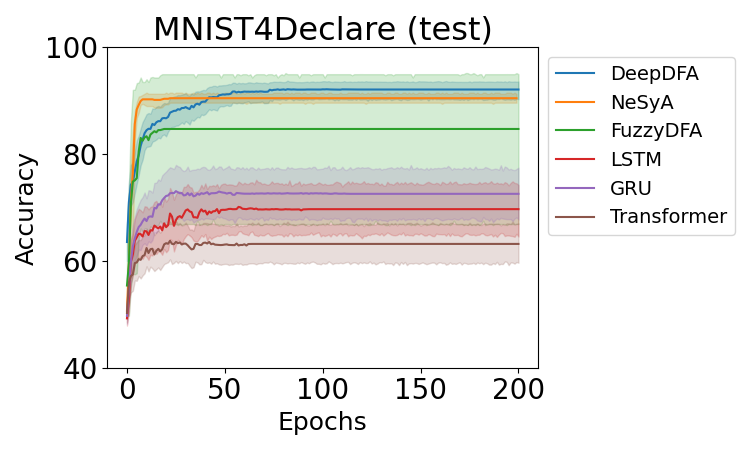}
    }
\subfigure[]{
    \includegraphics[width=0.32\textwidth]{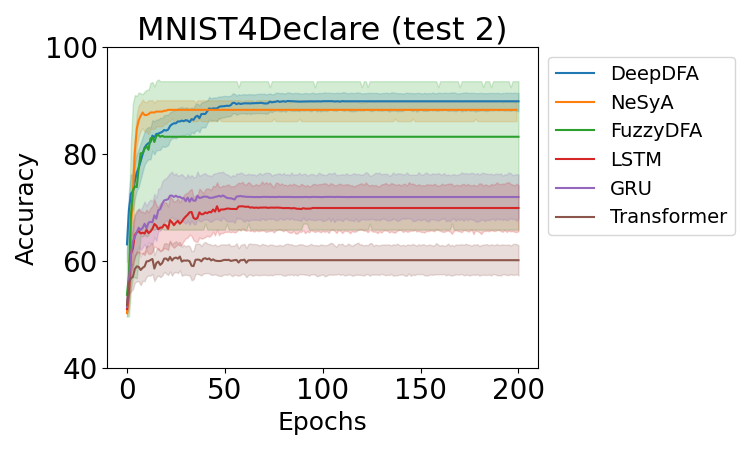}
    }

    \caption{Accuracy obtained in the image sequence classification task by the various approaches on the MNIST4Declare dataset. (a) training accuracy. (b-c) accuracy on test sequences respectively two (b) and three (c) times longer than train sequences.}\label{fig:sym_ground_results}
\end{figure*}
\subsubsection{Results on MNIST dataset}
We compare our neuro-symbolic approach with both classical supervised deep learning methods—using neural networks specialized for sequential data, such as LSTMs, GRUs, and Transformers—and state-of-the-art NeSy systems that incorporate temporal knowledge, such as FuzzyDFA \cite{Umili2023KR} and NeSyA \cite{NeSyA}.

For each approach and formula, we perform 10 runs with different random seeds and retain the best 7 runs based on training accuracy.
In every experiment, we report sequence classification accuracy, defined as the proportion of sequences correctly classified by the model.

To ensure a fair comparison, we use the same CNN-based symbol grounder across all experiments. For the LSTM, GRU, and Transformer baselines, we test two-layer models with either 24 or 64 hidden units. Interestingly, the smaller models (with 24 hidden units) consistently achieved better results, and we report those in the comparison.

Figure~\ref{fig:sym_ground_results} presents the mean results across 20 different Declare formulas. In all plots, the solid line represents the mean accuracy, while the shaded area indicates the standard deviation.

As the figure clearly shows, DeepDFA outperforms both purely neural and neuro-symbolic baselines. The purely neural models (LSTM, GRU, Transformer) rely solely on data, while the neuro-symbolic models (DeepDFA, NeSyA, FuzzyDFA) leverage both data and prior knowledge.
In particular, DeepDFA significantly outperforms FuzzyDFA on all datasets and slightly surpasses NeSyA on both test datasets—demonstrating state-of-the-art performance.

\subsubsection{Results on CAVIAR data}

For the experiment on the CAVIAR data, we use a CNN to classify each bounding box into one of four possible classes. The grounding of the symbol $close(p_1, p_2)$ is obtained by checking whether the distance between the two bounding boxes is smaller than a fixed threshold.
The probabilities of the nine short-term symbols are then combined to compute the probability of each of the 32 automaton symbols, according to the semantics shown in Figure~\ref{fig:caviar}(c). For example, the probability of symbol~0 is computed as the product of the probability that the first person is \emph{inactive}, the probability that the second person is \emph{inactive}, and $1 - close(p_1, p_2)$.
These probabilities form the input to DeepDFA, which predicts a probability distribution over the long-term events. The system is trained on sequences labeled at each time step with the corresponding long-term event, and performance is evaluated using the F1 score.

As a neurosymbolic baseline, we evaluate NeSyA under the same experimental setting, using the same CNN and the same prior knowledge. For the neural baselines, we use the same CNN but remove the final linear classification layer. The embedding vector extracted from the CNN is concatenated with the distance between the two bounding boxes, so that the neural baselines exploit the same information used by the neurosymbolic approaches. The resulting feature vector is subsequently processed by either an LSTM or a Transformer.

We report the results in Table~\ref{tab:learning_rate_results}. We test three different learning rates and, for each of them, report the mean and standard deviation of the F1 score over ten runs. The table also reports the number of trainable parameters for each method.
The performance of all methods varies noticeably across different learning rates. However, when considering the best test performance across learning rates (highlighted in the table), it becomes clear that the neurosymbolic approaches outperform the purely neural baselines. NeSyA and DeepDFA achieve comparable performance, with DeepDFA reaching the same best mean F1 score of 0.85 but with a smaller standard deviation. In contrast, the neural baselines do not exceed an F1 score of 0.75.

Regarding the number of parameters, both DeepDFA and NeSyA train only the CNN and therefore have significantly fewer trainable parameters than the other approaches.
\begin{table}[t!]
\centering
\small
\setlength{\tabcolsep}{3pt}
\resizebox{\columnwidth}{!}{%
\begin{tabular}{l c cc cc cc}
\hline
 &  & \multicolumn{6}{c}{\textbf{Learning rate}} \\
\cline{3-8}
\textbf{Model} & \textbf{\#Params}
& \multicolumn{2}{c}{0.001}
& \multicolumn{2}{c}{0.0001}
& \multicolumn{2}{c}{0.00001} \\
\cline{3-8}
 &  & Train & Test & Train & Test & Train & Test \\
\hline
DeepDFA & 258884
& $0.62 \pm 0.11$ & $0.53 \pm 0.16$
& $0.86 \pm 0.01$ & $\underline{\mathbf{0.85 \pm 0.11}}$
& $0.85 \pm 0.01$ & $0.73 \pm 0.18$ \\
NeSyA & 258884
& $0.81 \pm 0.13$ & $0.60 \pm 0.18$
& $0.87 \pm 0.03$ & $\mathbf{0.85 \pm 0.20}$
& $0.86 \pm 0.10$ & $0.81 \pm 0.18$ \\
CNN-LSTM & 399683
& $0.70 \pm 0.23$ & $\mathbf{0.56 \pm 0.21}$
& $0.84 \pm 0.08$ & $0.35 \pm 0.19$
& $0.17 \pm 0.05$ & $0.15 \pm 0.06$ \\
CNN-Transformer & 2659767
& $0.72 \pm 0.26$ & $0.40 \pm 0.10$
& $1.00 \pm 0.00$ & $0.68 \pm 0.16$
& $0.97 \pm 0.02$ & $\mathbf{0.78 \pm 0.14}$ \\
\hline
\end{tabular}%
}
\caption{Results for the CAVIAR dataset. Performance is averaged over 10 random seeds. The metric reported is macro F1 score. We present results for 3 different learning rates. For all systems training is stopped by monitoring the training loss with a patience of 10 epochs. Best test results for each method are in bold, while the best test performance across all setting is underlined.}
\label{tab:learning_rate_results}
\end{table}

\subsection{Reinforcement Learning}
For the second experiment we test the effectiveness of using DeepDFA for informing RL agents with the symbolic representation of their task by combining SSSG and RL, using the algorithm described in \ref{sec:integration_with_RL}. The code to reproduce these experiments is available at \url{https://github.com/KRLGroup/NeuralRewardMachines}

\begin{figure*}
        \centering
\subfigure[]{
    \includegraphics[width=0.23\textwidth]{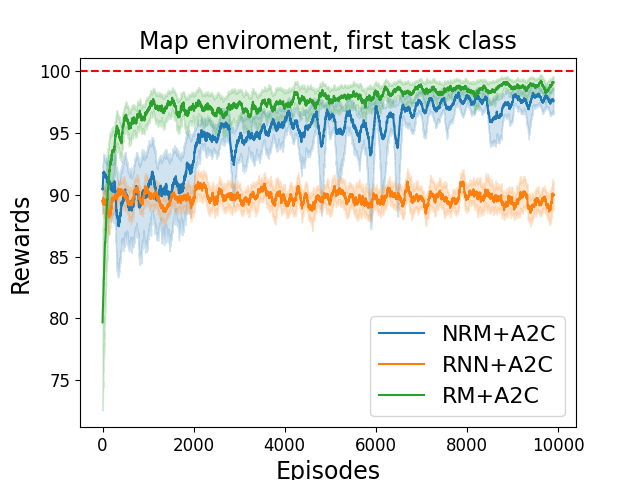}
    }
\subfigure[]{
    \includegraphics[width=0.23\textwidth]{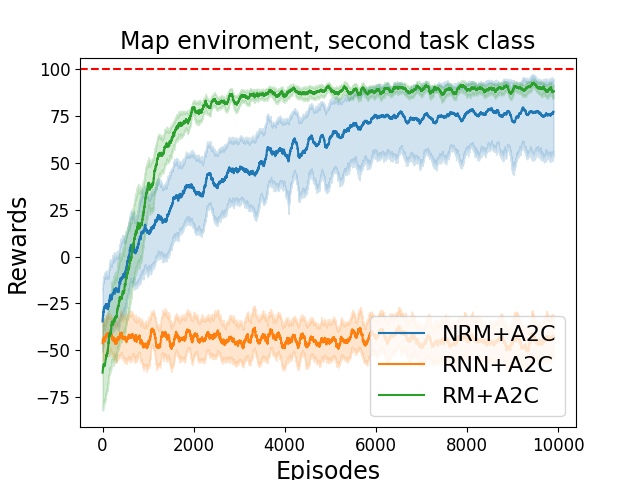}
    }
\subfigure[]{
    \includegraphics[width=0.23\textwidth]{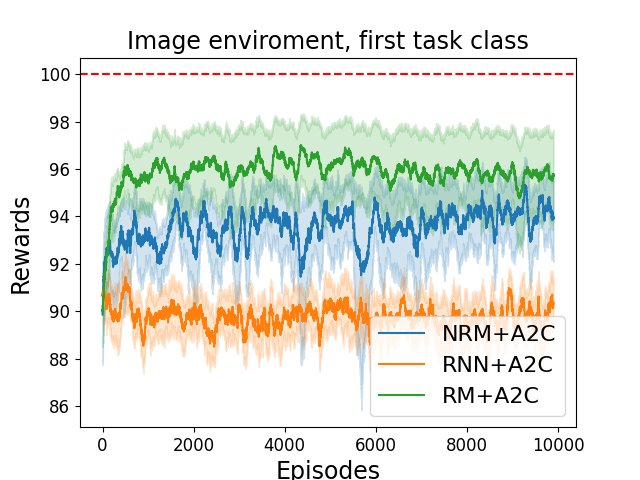}
}
\subfigure[]{
    \includegraphics[width=0.23\textwidth]{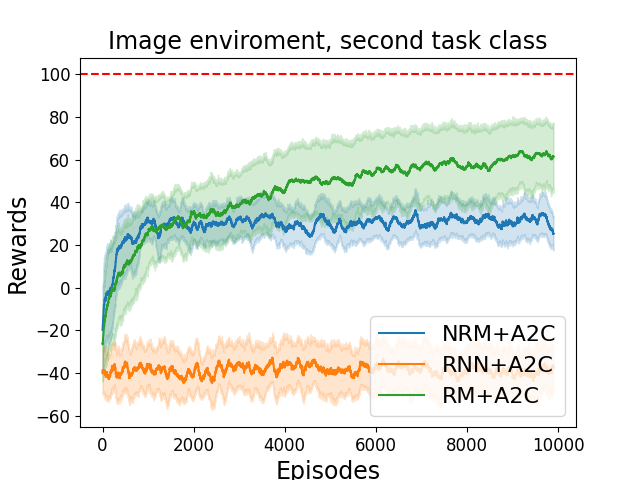}
}

\subfigure[]{
    \includegraphics[width=0.23\textwidth]{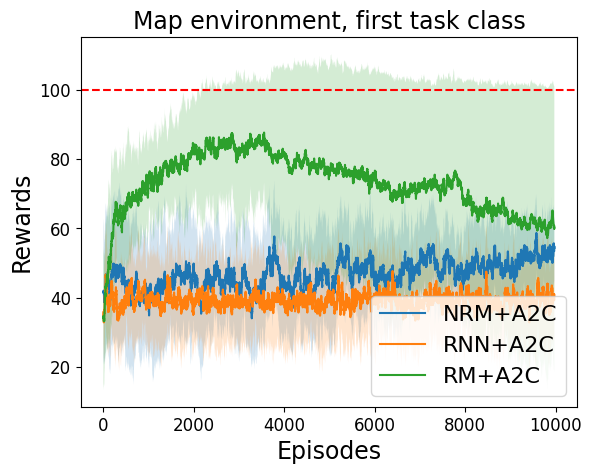}
    }
\subfigure[]{
    \includegraphics[width=0.23\textwidth]{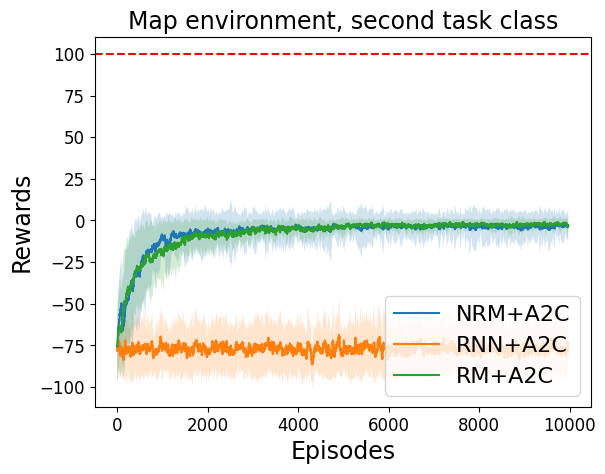}
    }
\subfigure[]{
    \includegraphics[width=0.23\textwidth]{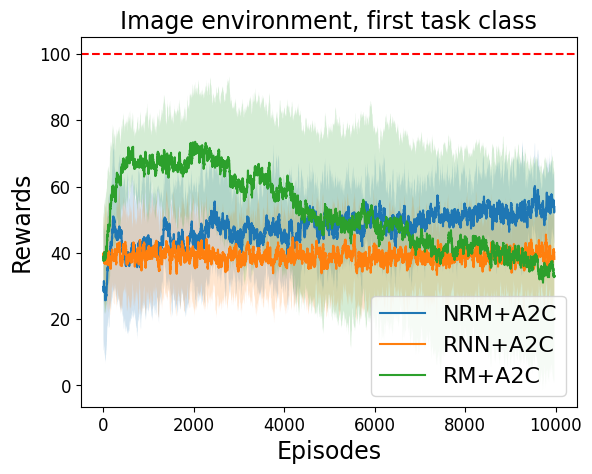}
}
\subfigure[]{
    \includegraphics[width=0.23\textwidth]{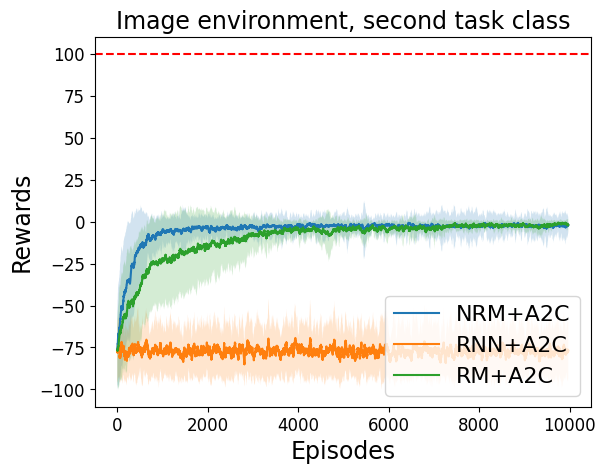}
}
    \caption{RL results under different configurations. First row (a,b,c,d): experiments with dense reward. Second row (e,f,g,h): sparse reward. First two columns(a,b,e,f): experiments in the map environment. Last two columns (c,d,g,h): experiments in the image environment. For each environment configuration we tested several task formulas, here results are aggregated by task class. Sequential tasks (first class) are shown in (a,c,e,f). Avoidance tasks (second class) are shown in (b,d,f,h)}
    \label{fig:results}
\end{figure*}
\subsubsection{Environment Design}
We test knowledge injection through DeepDFA on two types of environment inspired by the Minecraft videogame, similar to those considered by \cite{stochastic_reward_machines} and \cite{noisy_symbols_2022_shila} and described in the example of Section \ref{sec:nrm_and_RL}, in which we assume the labeling function is unknown.

\paragraph{States}
Based on this application we construct two environments that showcase different levels of difficulty in terms of symbol grounding:
(i) \textbf{The map environment}, in which the state is the 2D vector containing the x and y current agent location, (ii) \textbf{The image environment}, in which the state is an image of 64x64x3 pixels showing the agent in the grid, like the one shown in Figure \ref{fig:examples}(c).

\paragraph{Rewards}
We express the non-Markovian reward as an LTLf formula, that is then transformed into a DFA and, ultimately, into a Moore Machine. In the last process, each state $q$ is assigned a reward value, which is maximum in the automaton final states. 

We focus on patterns of formulas that are popular in non-Markovian RL \cite{RM_journal, LTL2action}, that we call here using the categorization given in \cite{ltl_patterns}.
\textbf{Visit} formulas: we denote as Visit($p_1$, $p_2$, ... , $p_n$) the LTLf formula F $p_1$ $\wedge$ F $p_2$ $\wedge$ ... $\wedge$ F $p_n$, expressing that the agent has to make true at least once each $p_i$ in the formula in any possible order.
\textbf{Sequenced Visit}: we denote with Seq\_Visit($p_1$, $p_2$, ... , $p_n$) the LTLf formula F($p_1 \land$ F($p_2 \land \ldots \land$ F($p_n$))), expressing that we want the symbol $p_{i+1}$ made true at least once \textit{after} the symbol $p_i$ has become True. \textbf{Global Avoidance}: we denote with Glob\_Avoid($p_1$, $p_2$, ... , $p_n$) the formula G ($\neg p_1$) $\wedge$ G ($ \neg p_2$) $\wedge$ ... $\wedge$ G ($ \neg p_n$), expressing that the agent must always avoid to make True the symbols $p_1$, $p_2$, ... , $p_n$. F and G are temporal operators called respectively `eventually' and `globally', we refer the reader to \cite{LTL} for the formal semantics of these operators that we explain here only intuitively. Based on these patterns of formulas we construct tasks of increasing difficulty that we aggregate in two classes. The \textbf{first class} is a set of tasks obtained as a conjunction of visit and sequenced visit formulas. The \textbf{second class} contains conjunctions of visit, sequenced visit and global avoidance formulas. We report in the appendix the formulas considered.

Reward sparsity strongly influences the ability of RL agents to find an optimal policy. Moreover, in our framework, the reward signal also serves as high-level supervision for the symbol grounder, as explained in Section~\ref{sec:rew_as_supervision}. Since it impacts both core modules of our framework—RL and SSSG—in different ways, we evaluate each environment under two reward settings: sparse and dense.
For the \textbf{sparse reward}, we adopt a three-value reward function \cite{LTL2action}, which assigns a positive value to final states, a negative value to failure states (i.e., states from which it is no longer possible to reach a final state), and zero to all other states. For the \textbf{dense reward}, we instead use a potential-based reward based on the distance in the automaton from the nearest final state, as illustrated in Figure~\ref{fig:examples}(d). In both cases, reward values are scaled so that the maximum cumulative reward is always 100, and the minimum reward is 0 or -100 for sequential and avoidance tasks, respectively.

\subsubsection{Comparisons} 
Given the absence of alternative methods in the literature for leveraging ungrounded prior knowledge in non-Markovian RL, we compare our \textbf{Neurosymbolic Reward Machine} (\textbf{NRM}) implemented with DeepDFA with the two most popular baselines in this domain: RNNs and RMs. However, it is important to note that the three methods assume different levels of knowledge.
Specifically, RM's performance serves as an upper bound for NRMs, as in the optimal scenario, where NRMs precisely learn the ground-truth SG function (which RMs possess from the outset), 
they become equivalent to RMs.
We use Advantage Actor-Critic (A2C) \cite{mnih2016asynchronous} as RL algorithm for all the methods. Hence we denote the methods (and which information they can exploit for learning): \textbf{RNN+A2C} (rewards) the baseline based on RNNs,  \textbf{NRM+A2C} (rewards+symbolic task) our method, and \textbf{RM+A2C} (rewards+symbolic task+grounding) the upper bound obtained with RMs.

\subsubsection{RL Results}
In Figure~\ref{fig:results}, we report the training cumulative rewards achieved by the three methods under different environment settings, varying the reward type (sparse vs.\ dense), the environment state representation (map vs.\ image), and the task class (class~1 vs.\ class~2). For each configuration and method, we perform 10 runs with different random seeds.

All methods share the same hyperparameter settings for A2C and the neural networks used for the policy, the value, and the feature extraction (only used in the image environment), which we detail in the appendix along with information on the network used as the grounder in the two environments.

Figure \ref{fig:results} clearly shows that the performance of NRMs lies between that of RNNs and RMs, and in many cases closely matches that of RMs, despite NRMs relying on less prior knowledge. In particular, under the dense reward setting (first row), RMs always achieve the maximum cumulative reward, while NRMs converge to almost the same reward values, especially in the map-based environment. Overall, RNNs consistently perform worse than the other methods, quickly converging to suboptimal local maxima, a behavior that is particularly evident in more challenging temporal tasks (class~2).

Under the sparse reward setting, the performance of all three methods degrades. Here, we observe two markedly different behaviors depending on the task class. For class~1 tasks, RMs converge to a suboptimal reward level (approximately 70--80) and then begin to exhibit forgetting, leading to a drop in performance. In contrast, NRMs and RNNs converge to even lower reward values, with NRMs performing slightly better than RNNs. This behavior can be attributed to the fact that NRMs receive insufficient feedback to correctly ground the symbols; consequently, the injected knowledge provides only limited benefits, as it cannot be effectively grounded in the observations.

For class~2 tasks, however, RMs and NRMs exhibit similar behavior—surprisingly, in the image-based environment, NRMs even outperform RMs. Both methods converge to a cumulative reward of 0, which corresponds to agents that learn to avoid forbidden symbols but fail to reach the target symbols. Symbols to avoid are easier to ground because they immediately produce negative rewards and are therefore less affected by reasoning shortcuts \cite{resoning_shrotcuts_survey}, making them easier to recognize. NRMs are likely able to identify and exploit this knowledge to avoid negative outcomes, but they struggle to recognize the sequence of symbols that must be collected in order (as also observed for class~1 tasks). As a result, they cannot fully leverage the injected knowledge to learn how to reach the goal states.

Interestingly, the performance of RMs shows that even with perfectly grounded symbols, the agent still fails to complete the required sequence. This occurs because the reward is too sparse, and with a random policy, the agent rarely completes the task. Extremely sparse reward settings are typically addressed using advanced exploration strategies, which are beyond the scope of this paper. Our primary objective is instead to investigate the effect of reward sparsity on symbol grounding. From this perspective, the observed results are consistent with our expectations, as already discussed in Section~\ref{sec:rew_as_supervision}, and similar to those obtained with perfect knowledge, therefore they are satisfying.

In summary, NRMs consistently outperform pure DRL baselines without knowledge injection. The largest benefits emerge when the feedback on the automaton state is sufficiently dense to allow effective symbol grounding in the observations. This occurs both in dense reward settings and in sparse reward settings for class~2 tasks. In all these scenarios, the performance of our neurosymbolic approach is comparable to that of RMs, despite relying on less prior knowledge about the task. These results demonstrate that DeepDFA can be successfully applied within a reinforcement learning setting.

\section{Conclusions and Future Work}
In conclusion, we introduce DeepDFA, a novel framework designed to leverage high-level logical knowledge—specifically in the form of deterministic finite automata (DFAs) or Moore machines—in a variety of non-symbolic sequential applications, ranging from classification to reinforcement learning. The core idea is to bridge the gap between subsymbolic perception and symbolic reasoning using a differentiable logic layer grounded in the theory of Probabilistic Finite Automata (PFA). This layer can be either initialized with temporal prior knowledge or trained directly from data, making it a flexible tool for both knowledge injection and temporal pattern learning.

This paper focuses primarily on knowledge integration, showing that prior symbolic knowledge can significantly improve performance and data-efficiency in tasks where temporal logic plays a central role. Across both static sequence classification and non-Markovian RL benchmarks, DeepDFA demonstrates strong performance, consistently outperforming purely neural models and state-of-the-art neuro-symbolic approaches.

Importantly, DeepDFA also advances the ongoing conversation around symbol grounding in temporal domains. Unlike most previous works that address grounding in static or offline settings, our approach allows grounding to be learned jointly with the RL policy in dynamic environments. This makes DeepDFA particularly relevant to real-world AI systems where the perception pipeline must be learned from raw data, but the process of observing such data is constrained by a dynamic environment and how well we know to explore it.


Future work will expand DeepDFA in two key directions. First, we aim to investigate automata induction techniques under the DeepDFA framework, allowing the system to learn temporal rules from both symbolic and raw sequences without relying on prior knowledge. Second, we plan to explore integration with autoregressive sequence generation models, enabling constrained sequence generation in domains like natural language generation and suffix prediction in BPM \cite{umili_ltlf_bpm}.



\section*{Acknowledgements}
This work has been partially supported by PNRR MUR project PE0000013-FAIR.

The work of Francesco Argenziano was carried out when he was enrolled in the Italian National Doctorate on Artificial Intelligence run by Sapienza University of Rome.

\bibliographystyle{unsrtnat}
\bibliography{references}

\end{document}